\documentclass[11pt]{article}

\usepackage[preprint]{acl}

\usepackage{times}
\usepackage{latexsym}

\usepackage[T1]{fontenc}

\usepackage[utf8]{inputenc}

\usepackage{microtype}

\usepackage{inconsolata}

\usepackage{graphicx}

\usepackage{booktabs}
\usepackage{multirow}
\usepackage{array}
\usepackage{float}
\usepackage{placeins}
\usepackage{stfloats}
\usepackage{amsmath}
\usepackage{amssymb}
\usepackage{color}

\usepackage[english]{babel}

\title{Evaluating LLM-Generated Preference Distributions}

\author{Fan Huang \\
  Indiana University Bloomington \\
  Bloomington, IN, USA \\
  \texttt{fanhuan@iu.edu} \\\And
  Minsuk Kim \\
  Massachusetts Institute of Technology \\
  Cambridge, MA, USA \\
  \texttt{dan418@mit.edu} \\ \AND
  C. Tyler Diggans \\
  Root Dynamix LLC \\
  Sewickley, PA, USA \\
  \texttt{tyler@rootdynamix.com} \\ \And
  Filippo Radicchi \\
  Indiana University Bloomington \\
  Bloomington, IN, USA \\
  \texttt{filiradi@iu.edu}
  }

\begin{document}
\maketitle

\begin{abstract}
Large Language Models (LLMs) are increasingly used as probabilistic generators for simulation, synthetic data generation, and decision support in settings where real-world data are unavailable. Yet, the structure and reliability of the distributions they produce remain understudied. Here, we systematically analyze LLM-generated distributions of preferences for air travel, restaurants, and consumer products.
Encouragingly, all models considered in our analysis exhibit self-coherence, with the most probable outcomes stabilizing rapidly under repeated sampling. At the same time, we observe substantial discordance across both model families and scales, with little consensus even among their most probable outcomes.
These patterns hold across nine open-weight models, three choice domains, and show robustness under temperature changes, greedy decoding, and perturbations of prompt and ordering.
Our findings indicate that outcomes are influenced more by the choice of model than by the wording of the prompt, challenging the common assumption that sufficiently capable LLMs produce similar preference distributions when used as stand-ins for survey respondents.
\end{abstract}

\section{Introduction}

There is a growing trend in relying on
Large Language Models (LLMs)
as probabilistic generators for
synthetic data creation \citep{argyle2023out}, agent-based simulation \citep{park2023generative}, opinion modeling \citep{santurkar2023whose}, and survey simulation \citep{tjuatja2024llms}. In a typical use case, an LLM is queried repeatedly under structured prompts, and its output is interpreted as a draw from some underlying, meaningful distribution.
This latent distribution being sampled is often referred as an \emph{implicit world prior}, that is, a reproducible mapping from a natural-language scenario description to a probability distribution over a discrete outcome set, induced by pre-training rather than by explicit calibration. Unlike a full ``world model'' \citep{ha2018world}, which implies an internal simulator for planning, an implicit world prior is defined operationally through externally observable distributional behavior.  These theorized distributions would only be useful for downstream simulation when they are stable under repeated sampling, internally structured rather than degenerate, and responsive in a bounded way to controlled prompt perturbations.

Recent work supports the plausibility of such priors: LLMs encode geographic and temporal regularities \citep{gurnee2024language}, can be repurposed as world models that supply state predictions to an external planning algorithm \citep{hao2023reasoning}, and can be characterized as models of the agents that produced their training data \citep{andreas2022agent}. Building on this, a growing body of work treats LLMs as probabilistic stand-ins for human populations and as agents in social simulation, while also documenting their sensitivity to prompt design; we review this literature in Section~\ref{sec:related}.

Despite this growing reliance on LLM-derived distributions, their structural properties remain underexplored. Existing evaluations largely target the factual accuracy \citep{lin2022truthfulqa}, confidence calibration, or aggregate benchmark performance \citep{hendrycks2021measuring} of single outputs, rather than the entire output distribution under repeated sampling, the differential effect of model scale on the latent prior, or the magnitude of prompt-induced shifts in distributional terms. Recent work names prompt sensitivity and the confounded state of LLM evaluation among the foundational open challenges in assuring LLM behavior \citep{anwar2024foundational}.

In this paper, we focus our attention on LLM-generated preference distributions for air
travel, restaurants, and consumer products.
We ask four questions: (i) Under repeated sampling at a fixed prompt, do LLM distributions converge, and, if so, which parts (high-probability modes or low-probability tails) stabilize first? (ii) Do different model sizes within the same family encode the same underlying prior? (iii) How sensitive are these distributions to demographic, contextual, and seasonal prompt perturbations? and (iv) Do the observed patterns generalize across choice domains and model families?

To answer these questions, we introduce a framework for evaluating LLMs' implicit world priors by prompting them to output explicit probability distributions over discrete choice sets, applied to nine open-weight models across three families (Llama, Qwen, Mistral) and three domains (travel, restaurants, products). As a desired outcome,  we observe, irrespective of the domain of application and the LLM's model family considered, a consistent behavior characterized by a relatively quick consolidation of the modes of the distributions.
On the other hand, we also find that
models from different families generate rather different distributions, and that even within the same family, there is
a strong scale discordance:
for Llama models (8B to 405B), different scales encode structurally different priors over the same choice set, with pairwise Jensen--Shannon (JS) divergence in the range 0.36 to 0.50 and top-5 agreement below 0.13, larger than any bounded shift induced by prompt perturbations (JS in the range 0.13 to 0.33); this discordance persists, though more mildly, across the Qwen and Mistral families.
All findings are validated under temperature ablation, greedy decoding, paraphrase and permutation robustness, and choice-set size variation; code and processed data will be made publicly available.
Our results question the implicit assumption in silicon-sampling work \citep{argyle2023out, park2023generative} that a single fixed prior can be elicited from any LLM.

\section{Related Work}
\label{sec:related}

\subsection{LLMs for Simulation and Prediction}

Recent work has explored LLMs as simulators of human behavior \citep{park2023generative}. \citet{argyle2023out} show that conditioning on socio-demographic backstories yields response distributions that correlate with those of the matching human subgroups, whereas \citet{santurkar2023whose} report substantial misalignment between the opinion distributions of current LLMs and those of US demographic groups, a gap that persists even when the models are explicitly steered toward a group. \citet{tjuatja2024llms} similarly found that LLMs generally fail to reflect human-like response biases under question modifications known to elicit those biases in humans, and \citet{anwar2024foundational} identify prompt sensitivity and confounded evaluation among the foundational open challenges in assessing LLM behavior. However, the distributional properties of LLM outputs themselves (stability, convergence, and model sensitivity) remain largely uncharacterized at the level of full output distributions.

\subsection{Probabilistic Calibration and Uncertainty}

Research on neural network calibration has long examined whether model confidence aligns with accuracy \citep{guo2017calibration}. In the context of LLMs, \citet{jiang2021how} investigated calibration for question answering, while \citet{kadavath2022language} showed that language models can often assess whether they know the answer to a question. More recent work has developed methods for quantifying semantic uncertainty in generated text \citep{kuhn2023semantic} and explored whether LLMs can verbally express their uncertainty \citep{xiong2024can}. While these studies focus on calibration of single-point predictions or confidence scores, our work analyzes full probability distributions over outcome spaces, providing a more complete characterization of LLM distributional behavior.

\subsection{Prompt Sensitivity and Robustness}

Studies have documented that LLM outputs can be highly sensitive to prompt phrasing, including the order of few-shot examples \citep{lu2022fantastically} and biases toward majority labels or recent examples \citep{zhao2021calibrate}. \citet{perez2021true} demonstrated that prompt selection can substantially affect few-shot performance, while \citet{webson2022prompt} questioned whether models truly understand prompt semantics or rely on superficial patterns. Prompting strategies such as chain-of-thought \citep{wei2022chain} have been shown to elicit more reliable reasoning, yet the sensitivity of outputs to prompt variations remains a fundamental challenge.

\subsection{Mode Collapse and Numeric Calibration}

A related line of work examines \emph{mode collapse}, where generation diversity degrades during fine-tuning \citep{janus2022mode,omahony2024attributing}.
\citet{zhou2024reluctance} showed that LLMs are reluctant to express uncertainty, relevant to our verbalized elicitation design. \citet{lovering2025calibration} demonstrated poor numeric calibration in language models.
\citet{lum2025bias} advocated evaluation beyond isolated test cases, a perspective aligned with our distribution-level framework.

\section{Methods}

We prompt LLMs to generate explicit probability distributions over large discrete choice sets, enabling direct comparison across repeated queries. Throughout the paper, our primary domain is \emph{air-travel destination choice}: each query elicits a distribution over a fixed choice set of 300 contiguous-US commercial airports, adopted from prior work on the US air transportation network \citep{kim2025modeling}. The framework itself is domain-agnostic; we additionally study restaurant choice (Yelp) and product purchase (Amazon), but use these two domains only to \emph{cross-validate} that our findings generalize beyond air travel (Section~\ref{sec:cross_domain_results}).

\subsection{Problem Formulation}

We treat LLMs as
\emph{implicit world priors} that manifest as probability distributions over discrete outcome spaces conditioned on natural language prompts.
Formally, given a prompt $x$ describing a contextual scenario (e.g., departure airport, traveler persona), an LLM induces a probability distribution over outcomes $y \in \mathcal{Y}$:
\(p(y \mid x)\).
In our setting, $\mathcal{Y}$ corresponds to a finite set of candidate destinations. Rather than treating LLM outputs as single-point predictions, we explicitly analyze the structure, stability, and sensitivity of $p(y \mid x)$ under repeated sampling and controlled prompt perturbations.

\subsection{Elicitation Method Validation}
\label{sec:elicitation_validation}

We adopt \emph{verbalized distribution elicitation}: prompting the LLM to output an explicit probability distribution over all options in a single query. \citet{hu2023prompting} argue that direct probability readout via logits is more faithful than metalinguistic prompting; however, both they and \citet{lovering2025calibration} examine \emph{calibration-to-ground-truth}, whereas our analysis targets \emph{reproducibility} and \emph{structural consistency}, properties for which verbalized elicitation is sufficient. The two natural alternatives, logit inspection and repeated outcome sampling, are both infeasible at our experimental scale (see Appendix~\ref{sec:appendix_elicitation} for the full comparison and pilot study).

\subsection{Iterative Sampling Protocol}

For each prompt $x$,
we invoke the LLM under identical settings $R$ times to produce a set of sampled distributions $\{ p^{(r)}(y \mid x) \}_{r=1}^{R}$,
with mean $p(y|x) = \frac{1}{R} \sum_{r=1}^R p^{(r)}(y|x)$.
Each invocation is an independent API call with no shared state, conversation history, or seed dependency; the index $r$ is an arbitrary label, not a temporal ordering.
Following our empirical observations at temperature $T=0.2$ (validated in Section~\ref{sec:temperature_sensitivity}), each model is queried for
a maximum number of samples to $R_{\max} = 20$, which is sufficient to observe
convergence
of the mean sample distribution
across all model sizes at this temperature setting, while keeping the number  of the model's invocations reasonable.
Cumulative metrics computed over the sample pool are invariant to sample ordering by construction; we separately validate robustness to the ordering of the airport list within the prompt via a permutation analysis (Section~\ref{sec:statistical_robustness}).

\subsection{Mode and Tail Regions}
\label{subsec:mode_tail}
To characterize internal distributional dynamics, we partition each distribution's outcomes into high-probability modes and low-probability tails.
For a fixed non-negative integer $k$, order the outcomes by decreasing probability, $p(y_{(1)} \mid x) \ge p(y_{(2)} \mid x) \ge \cdots \ge p(y_{(|\mathcal{Y}|)} \mid x)$ (ties broken arbitrarily), and define the mode set as the $k$ most probable outcomes, \(\mathcal{M}_k(p) = \{y_{(1)}, \ldots, y_{(k)}\}\).
The tail is defined as the complement \(\mathcal{T}_k(p) = \mathcal{Y} \setminus \mathcal{M}_k(p)\).
We note that this mode--tail distinction is temperature dependent.
Our use of low temperature ($T=0.2$) is designed to minimize this effect; see Section~\ref{sec:temperature_sensitivity} for a systematic analysis of temperature's influence on mode--tail separation.
This partition enables us to separately analyze the stability of dominant semantic beliefs (modes) and exploratory variability in low-probability outcomes (tails).

In our experiments, we set $k = 5$, which is empirically grounded: across 300 departure contexts, the natural mode--tail boundary (largest consecutive probability gap) falls at rank $\leq 5$ in 99\% of distributions. A sensitivity analysis over $k \in \{3, 5, 7, 10, 15\}$ confirms that the tail-dominated ordering and the level of mode agreement are preserved at every cutoff $k \geq 5$ (Table~\ref{tab:k_sensitivity}; see Appendix~\ref{sec:appendix_k_justification} for the full justification).

\subsection{Evaluation Metrics}
\label{sec:evaluation_metrics}

We characterize distributional behavior using the following families of metrics.\footnote{We use ``distributional consistency'' to denote reproducibility under repeated sampling, distinct from ``distributional stability'' in dynamical systems and from out-of-distribution generalization \citep{liu2024need}.}

\paragraph{Distributional Consistency.} We quantify this with Jensen--Shannon (JS) divergence \citep{lin1991divergence} for overall similarity, Jaccard@$k$ for top-$k$ mode overlap, and Spearman correlation for rank consistency.

\paragraph{Tail Dynamics.} We track support growth, Shannon entropy, and tail mass (probability outside the top-$k$ modes). Let $\mathcal{S}_r \subseteq \mathcal{Y}$ denote the \emph{support} of sample $r$ (the outcomes assigned non-zero probability, a subset of the fixed choice set $\mathcal{Y}$), and $|\mathcal{S}_r|$ its size. \emph{Support growth} is the expansion of this realized support as more independent samples are collected. Let $\mathcal{C}_R = \bigcup_{r=1}^{R} \mathcal{S}_r$ be the \emph{cumulative support}: the set of distinct outcomes observed across the first $R$ samples (the union of their per-sample supports), reported as Unique@$R$ in results tables. The \emph{growth percentage} is $\text{Growth} = (|\mathcal{C}_R| - |\mathcal{C}_1|) / |\mathcal{C}_1| \times 100\%$, with $\mathcal{C}_1 = \mathcal{S}_1$ the support of a single sample. Because $\mathcal{C}_R$ is a union of sets, it does not depend on sample order; the growth percentage depends only mildly on which sample is taken first (through $\mathcal{C}_1$), and the permutation analysis in Section~\ref{sec:statistical_robustness}, which shuffles the airport list within the prompt, confirms robustness (CV = 13.2\% across 15 random orderings). We summarize a distribution's overall breadth by its \emph{final support}: the number of distinct outcomes with non-negligible probability ($>0.1\%$, i.e.\ $\epsilon = 0.001$) in the converged distribution, pooled over all $R$ samples.

\paragraph{Cross-Condition Analysis.} We compute JS divergence and Jaccard@$k$ between distributions under different prompt conditions (persona, demographics, location).

\paragraph{Convergence.} The \emph{stabilization point} $R^*$ is the smallest sample count beyond which the top-$k$ mode set $\mathcal{M}_k$ of the distribution pooled over the first $R$ samples becomes, and remains, identical as $R$ grows (top-$k$ Jaccard $=1$ between successive cumulative sets). We locate this point by thresholding the top-$k$ Jaccard index; the threshold is a robustness detail rather than a tuned parameter, since any value in a wide band just below $1$ recovers the same $R^*$ (Appendix~\ref{sec:appendix_stab_threshold}). Separately, results tables report a coarser, support-based stabilization signal, Stable@: the first sample from which the cumulative support size changes by at most two outcomes over three successive samples. Because it is a support-level criterion rather than a top-$k$ set-identity one, Stable@ can settle earlier than $R^*$ and is reported only as a descriptive complement.

\paragraph{Region-Wise JS Divergence.} To analyze distributional variability, we compute JS divergence on the full distribution ($\mathrm{JS}_{\mathrm{total}}$) and separately on renormalized mode ($\mathrm{JS}_{\mathrm{mode}}$) and tail ($\mathrm{JS}_{\mathrm{tail}}$) regions. These provide complementary views: $\mathrm{JS}_{\mathrm{total}}$ captures aggregate distributional change, while $\mathrm{JS}_{\mathrm{mode}}$ and $\mathrm{JS}_{\mathrm{tail}}$ isolate whether variability arises from core semantic beliefs or peripheral exploration. Note that they do \emph{not} form an additive decomposition.

For scale discordance analysis, we compare distributions across model sizes using both JS divergence and agreement@$k$ (proportion of contexts with identical top-$k$ sets). The joint interpretation of these metrics distinguishes scale refinement (consistent beliefs across sizes) from scale discordance (fundamentally different priors). Complete mathematical definitions are provided in Appendix~\ref{sec:appendix_metrics}.

\section{Experiments}
\label{sec:experimental_setup}

Our experimental approach is described in terms of our primary domain of air-travel destination choice for clarity, but additional experiments were carried out for cross-domain validation on restaurants (Yelp) and product choice (Amazon) separately, with results presented in Section~\ref{sec:cross_domain_results}. For these airline queries, we employ structured prompts providing LLMs with a traveler persona, a departure airport, 300 possible destination airports, and instructions to output preferences as probabilities. Figure~\ref{fig:pipeline} summarizes the end-to-end pipeline.
Personas are defined across 5 age groups and 4 education levels, yielding 20 unique combinations; the grid is unweighted, and the American Community Survey~\citep{acs2025} provides demographic background rather than population weights.
The 300-airport choice set is representative of the contiguous-US air transportation network, as considered in prior work~\citep{kim2025modeling};
the departure airport is also selected from this set. Output probabilities are eventually normalized for mathematical consistency. Models also occasionally return non-conforming outputs, for example airport names in non-standard formats or otherwise malformed responses (informally, ``hallucinations''), which we resolve to canonical three-letter airport codes where possible and discard otherwise; Appendices~\ref{sec:appendix_output_norm} and~\ref{sec:appendix_parsing} detail this process.

Experiments primarily use Llama-3.3-70B via the Together AI API\footnote{\url{https://api.together.ai/}}, with extension experiments spanning 9 models across three families (Llama, Qwen \citep{qwen2.5}, Mistral \citep{mistral7b,mixtral}) at small, medium, and large scales (Table~\ref{tab:model_config}). All experiments use the temperature $T=0.2$, validated across 10 temperature values and 6 experimental settings as producing consistent mode concentration and high ranking stability (Section~\ref{sec:temperature_sensitivity}; Appendix~\ref{sec:appendix_temperature}). All experiments were additionally replicated under greedy decoding ($T{=}0$) to confirm that findings reflect intrinsic model properties rather than sampling artifacts (Appendix~\ref{sec:appendix_greedy}). See Appendix~\ref{sec:appendix_experimental} for complete configuration details.

LLM outputs are generated as free-form JSON and parsed with post-hoc rectification (Appendix~\ref{sec:appendix_parsing}). While constrained decoding and structured output modes can guarantee well-formed responses, these features are not uniformly available across all model families and API providers tested in our study. To ensure that our findings generalize across LLMs rather than depending on provider-specific output enforcement, we adopt free-form generation as the common baseline for all models.

In total, we evaluate four departure airports (IND, FWA, ATL, ORD) under the persona, model, and temperature settings described above. The temperature ablation comprises 600 API calls (Appendix~\ref{sec:appendix_temperature}), the multi-source robustness analysis 345 (Appendix~\ref{sec:appendix_robustness}), and the cross-family extension 339 (Appendix~\ref{sec:appendix_crossfamily}). See Appendix~\ref{sec:appendix_datasets} for dataset descriptions.

\paragraph{Sample-size bootstrap protocol.}
\label{sec:bootstrap_protocol}
To study the convergence of the various metrics under repeated sampling, we use our $R_{\max} = 20$ total experiments to create random subsamples of size $R \in \{1, \ldots, R_{\max}\}$. Specifically,
we draw $R$ samples \emph{without replacement} from the pool, compute the metric on that subset, and repeat the draw $\min(100, \binom{R_{\max}}{R})$ times: whenever the number of unique size-$R$ subsets satisfies $\binom{R_{\max}}{R} \le 100$ we enumerate every unique subset exactly once,
otherwise we draw $100$ random unique subsets. We plot the mean together with the $[2.5, 97.5]$ percentile band as a $95\%$ confidence interval. Transition-based metrics (e.g., pairwise $\mathrm{JS}$ divergence between consecutive sample sizes) use a pool of 19 transitions following the same procedure. The CI band at $R=R_{\max}$ is
by definition equal to zero because each resample then equals the full pool.

This is a property of without-replacement subsampling rather than a convergence finding. The informative part of every such curve is therefore the small-$R$ region, where the band reflects the variability that would be observed with a small number of independent samples. We report the full curve to make this sample-size sensitivity explicit, but the right-hand endpoint should be read as a sanity check rather than as evidence of convergence.

\paragraph{Numerical precision in tables.}
For each numeric cell in the tables of results, we report the mean $\pm$ half-width of the $95\%$ bootstrap CI, computed by resampling the relevant unit of analysis (personas for cross-scale and cross-family comparisons; pairwise distribution comparisons for prompt perturbations). The displayed precision of each point estimate is rounded to match the precision of the bootstrap CI half-width (i.e., to one significant figure on the CI), so that the trailing digit of the mean is at most as fine as the uncertainty of the estimate. The same rounding rule applies to numerics quoted in the abstract, introduction, and results narrative. Single-context entries for which a bootstrap CI is not defined are flagged explicitly in the relevant caption.

\section{Results}
\label{sec:results}

We organize our findings around five empirical claims that characterize the structure and behavior of LLM-generated distributions as reflections of implicit world priors. Full probability matrices and per-question result summaries are provided in Appendix~\ref{sec:appendix_results}. The five claims are presented in turn: Section~\ref{sec:rapid_stabilization} shows that the modes of the distributions stabilize rapidly under repeated sampling; Section~\ref{sec:tail_dominated} shows that the residual post-convergence variability is tail-dominated; Section~\ref{subsec:scale_discordance} shows that model scale induces distributional discordance; Section~\ref{sec:prompt_perturbations} shows that prompt perturbations produce bounded, structured shifts; and Section~\ref{sec:cross_domain_results} shows that these patterns generalize across choice domains. Section~\ref{sec:robustness} then reports robustness checks that support all five claims.

\subsection{Rapid Stabilization
of Modes of the Distributions
}
\label{sec:rapid_stabilization}

Our primary finding is that LLM-generated distributions exhibit a consistent modes-first convergence pattern: high-probability outcomes are consistent across independent samples, while variability is concentrated in low-probability alternatives. Across contexts and model sizes, the cumulative destination support stabilizes within a few samples (Table~\ref{tab:convergence}). The dominant modes show high pairwise agreement: across all sample pairs, mean Jaccard@5 is 0.72 and 36 of the 74 sample pairs share an identical top-5 set (Jaccard@5 $=1.0$), with Spearman rank correlation exceeding 0.95 at stabilization points. The per-model variation in stabilization rate and unique-destination count, particularly the much later stabilization (sample 17) and higher diversity (100 vs.\ 32--54 destinations) of the 8B model, is discussed under scale discordance in Section~\ref{subsec:scale_discordance}. Full iteration heatmaps and per-context stability metrics are provided in Appendix~\ref{sec:appendix_convergence}.

\begin{table}[t]
\centering
\small
\begin{tabular}{@{}lccc@{}}
\toprule
\textbf{Context/Model} & \textbf{Unique@1} & \textbf{Unique@20} & \emph{Stable@} \\
\midrule
IND & 19 & 32 & Iter 3 \\
FWA & 16 & 42 & Iter 6 \\
ATL & 15 & 43 & Iter 5 \\
ORD & 18 & 34 & Iter 3 \\
\midrule
\multicolumn{4}{c}{\textit{Model Size Comparison (IND)}} \\
\midrule
Baseline (70B) & 19 & 32 & Iter 3 \\
Larger (405B) & 30 & 54 & Iter 4 \\
Smaller (8B) & 29 & 100 & Iter 17 \\
\bottomrule
\end{tabular}
\caption{Convergence results for travel demand (US airports). \emph{Unique@$R$}: cumulative support size, distinct destinations with non-zero probability pooled over the first $R$ samples (Section~\ref{sec:evaluation_metrics}). \emph{Stable@}: the first sample from which this cumulative support plateaus (changes by at most two destinations over three successive samples); a support-size stabilization signal, complementary to and distinct from the top-$k$ stabilization point $R^*$ (Section~\ref{sec:evaluation_metrics}).}
\label{tab:convergence}
\end{table}

\subsection{Post-Convergence Variability is Tail-Dominated}
\label{sec:tail_dominated}

Having established rapid mode stabilization, we next examine post-convergence variability. Our analysis reveals that nearly all remaining divergence arises from tail outcomes rather than from instability in core semantic beliefs.

The source of variability is localized by computing JS divergence separately for mode and tail regions. As samples accumulate, both regions contract while the tail consistently dominates: post-convergence (iterations 5--20), tail divergence remains above mode divergence (JS$_{\text{mode}}$ = 0.058--0.059 vs.\ JS$_{\text{tail}}$ = 0.124--0.132; a tail/mode ratio of 2.1--2.3$\times$, averaged across the four contexts; Table~\ref{tab:mode_tail_decomp}). This confirms that residual variability is driven by low-probability outcomes, but this gap is expected, as renormalizing the noisy tail outcomes turns their small absolute fluctuations into larger relative divergence (Appendix~\ref{sec:appendix_metrics}). Tail support size remains stable (17--20 unique destinations), with 36.7--40.2\% of probability mass outside the top-5 modes (per-airport means across the four contexts), indicating sustained uncertainty over non-modal alternatives.  The observed pattern is an empirical regularity and is indicative of a potential combination of a true tail of an implicit world prior with temperature-driven perturbations. Temperature modulates the magnitude of this variability but does not account for it: raising $T$ from $0.1$ to $1.0$ widens the tail systematically (Appendix~\ref{sec:appendix_temperature}), yet under greedy decoding ($T{=}0$) the tail contracts only from $32$ to $27$ unique destinations for IND while the modes-first pattern is unchanged (Appendix~\ref{sec:appendix_greedy}).

\begin{table}[t]
\centering
\small
\setlength{\tabcolsep}{3pt}
\begin{tabular}{@{}lcccc@{}}
\toprule
\textbf{Iter.} & $\mathrm{JS}_{\text{total}}$ & $\mathrm{JS}_{\text{mode}}$ & $\mathrm{JS}_{\text{tail}}$ & $\Delta_{\text{tail}-\text{mode}}$ \\
\midrule
1--3   & $0.12{\pm}0.06$ & $0.07{\pm}0.06$ & $0.20{\pm}0.07$ & $+0.13{\pm}0.03$ \\
3--5   & $0.19{\pm}0.09$ & $0.13{\pm}0.08$ & $0.31{\pm}0.11$ & $+0.18{\pm}0.05$ \\
5--10  & $0.08{\pm}0.03$ & $0.06{\pm}0.04$ & $0.12{\pm}0.04$ & $+0.06{\pm}0.03$ \\
10--20 & $0.09{\pm}0.03$ & $0.06{\pm}0.03$ & $0.13{\pm}0.04$ & $+0.07{\pm}0.04$ \\
\bottomrule
\end{tabular}
\caption{JS divergence by convergence phase on the full distribution, the renormalized top-5 mode region, and its complement (US airport travel demand). Cells are mean $\pm$ half-CI (\S\ref{sec:bootstrap_protocol}) over consecutive-sample comparisons pooled across the four contexts ($N = 12, 8, 20, 34$ by phase). $\Delta_{\text{tail}-\text{mode}}$, the paired difference, is the appropriate test of tail dominance since both regions share a sample pair: it excludes zero in every phase, whereas the marginal intervals overlap in three of the four.}
\label{tab:mode_tail_decomp}
\end{table}

\subsection{Distributional Discordance Across Scales and Families}
\label{subsec:scale_discordance}

Our most surprising finding is distributional discordance across models: LLMs encode structurally different latent world priors not only across model families but also across scales within a single family, exhibiting extremely low distributional agreement even under identical prompts and decoding parameters. This indicates that larger models do not simply refine smaller ones, nor do different families converge on a shared prior; each model encodes its own.

\begin{figure*}[t]
\centering
\includegraphics[width=\textwidth]{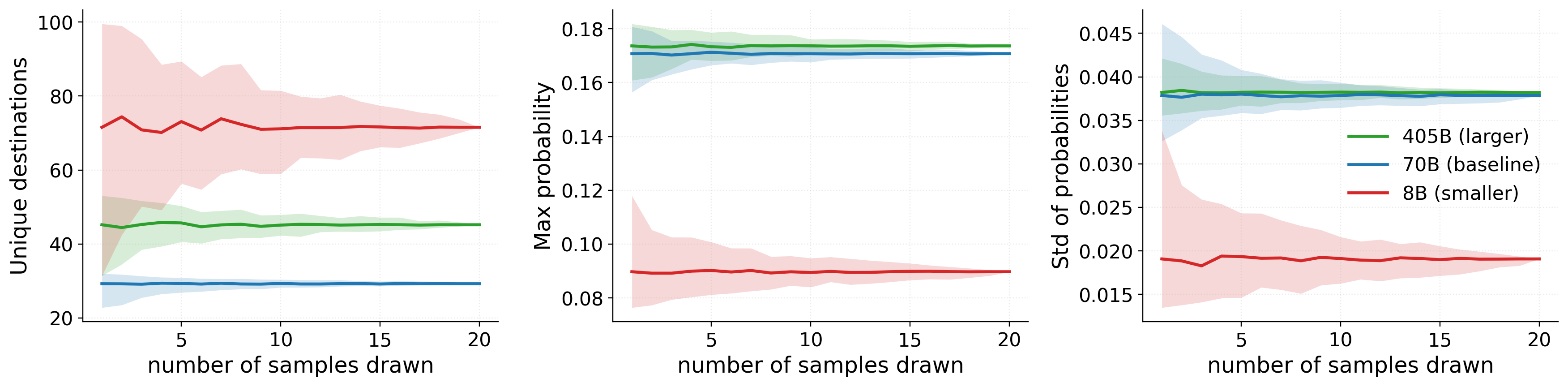}
\caption{Per-sample distributional signatures across three LLM sizes (Section~\ref{sec:bootstrap_protocol} resampling protocol; the CI band for the number of sampled distributions equal to $20$ is by definition zero because there is only one sample equal to the full pool, so the informative part of each curve is the small-$R$ region). (a)~Unique destinations: 8B (mean 71.6) is 1.58$\times$ the 405B (45.2) and 2.45$\times$ the 70B (29.3). (b)~Maximum destination probability: 70B and 405B concentrate 0.171 and 0.174, respectively; 8B is more diffuse (0.090). (c)~Std of probabilities: 70B and 405B at 0.038; 8B tighter (0.019). The three models occupy structurally different operating points, not refinements of one another: across the 20 personas no two share an identical 70B/405B top-5 (mean top-5 overlap $0.12$) and even the single dominant destination differs in most cases (top-1 match $0.30$; Table~\ref{tab:model_agreement}), so the larger model re-weights \emph{which} destinations dominate rather than sharpening the same modes.}
\label{fig:llm_convergence}
\end{figure*}

Table~\ref{tab:model_agreement} presents pairwise agreement metrics across three model sizes (8B, 70B, 405B). JS divergences are substantial (0.36--0.50, all $\pm$ 0.04--0.05), agreement@5 is extremely low (0.08--0.12), and top-1 match rates span only 5\%--30\% with substantial uncertainty given the persona sample size ($N = 20$). For comparison, within a single model the top-5 set of the pooled distribution is identical from the stabilization point onward (cumulative Jaccard@5 $=1.0$; Section~\ref{sec:evaluation_metrics}); the cross-scale agreement is thus an order of magnitude below typical within-model stability, meaning that swapping the model scale changes the dominant outcomes far more than re-sampling from a fixed model does. Figure~\ref{fig:llm_convergence} visualizes the per-sample distributional signature of each scale: the three models operate at clearly separated levels in unique-destination count, top destination probability, and probability spread, illustrating that the differences are structural rather than calibration-level. Further analysis is in Appendix~\ref{sec:appendix_models}, including the non-monotonic scaling of distributional diversity (Appendix~\ref{sec:appendix_nonmonotonic}) and per-scale stability metrics (Appendix~\ref{sec:appendix_stability}).

\begin{table}[t]
\centering
\small
\setlength{\tabcolsep}{5pt}
\begin{tabular}{@{}lccc@{}}
\toprule
\textbf{Model Pair} & \textbf{JS Div.} & \textbf{Agr.@5} & \textbf{Top1} \\
\midrule
70B vs 405B & $0.36 \pm 0.04$ & $0.12 \pm 0.05$ & $0.30 \pm 0.20$ \\
70B vs 8B   & $0.47 \pm 0.05$ & $0.08 \pm 0.04$ & $0.05 \pm 0.07$ \\
405B vs 8B  & $0.50 \pm 0.04$ & $0.08 \pm 0.03$ & $0.20 \pm 0.18$ \\
\bottomrule
\end{tabular}
\caption{Scale discordance metrics across three Llama sizes (US airport travel demand). Mean $\pm$ half-CI over 20 personas; see \S\ref{sec:bootstrap_protocol} for the precision convention.}
\label{tab:model_agreement}
\end{table}

\paragraph{Cross-Family Validation.}

To assess whether scale discordance is architecture-specific, we replicate our analysis across two additional model families: Qwen \citep{qwen2.5} and Mistral \citep{mistral7b,mixtral}. This extension spans nine models covering three training origins, three size tiers, and both dense and mixture-of-experts architectures. As shown in Table~\ref{tab:cross_family_discordance}, substantial scale discordance persists across all families: between comparable model sizes, JS divergence ranges from
0.14 to 0.28
for Qwen and Mistral, and from 0.36 to 0.50 for Llama. These two ranges are not measured on the same footing: the Llama figures are means over 20 personas, whereas the Qwen and Mistral figures are single-context point estimates. Computed single-context like the others, Llama spans 0.17 to 0.33, so once the protocol is matched the families are comparable in magnitude and the phenomenon is common to all three. While the degree of inter-scale agreement varies, for example, Qwen exhibits higher correlation at smaller scales than Llama; no family is free from this phenomenon. The per-scale distributional signatures for Qwen and Mistral (Figures~\ref{fig:qwen_convergence} and~\ref{fig:mistral_convergence}, Appendix~\ref{sec:appendix_crossfamily_signatures}; companion to Figure~\ref{fig:llm_convergence}) make this concrete: within each family, the scales occupy clearly separated operating points in unique-destination count, top-destination probability, and spread, rather than refinements of one another. In addition, all families exhibit the same modes-first convergence behavior: the cumulative support stabilizes within a few samples for every model (Qwen by iterations 3--6, Mistral by 2--5), consistent with the convergence dynamics observed in Llama, confirming that this pattern is architecture-independent.

\begin{table}[t]
\centering
\small
\begin{tabular}{@{}lccc@{}}
\toprule
\textbf{Family} & \textbf{S-M} & \textbf{M-L} & \textbf{S-L} \\
\midrule
Llama   & $0.47 \pm 0.05$ & $0.36 \pm 0.04$ & $0.50 \pm 0.04$ \\
Qwen    & 0.14$^\dagger$ & 0.15$^\dagger$ & 0.23$^\dagger$ \\
Mistral & 0.22$^\dagger$ & 0.14$^\dagger$ & 0.28$^\dagger$ \\
\bottomrule
\end{tabular}
\caption{Cross-family scale discordance (JS divergence; US airport travel demand). Llama: mean $\pm$ half-CI over 20 personas (same protocol as Table~\ref{tab:model_agreement}). $^\dagger$ single-context point estimate (IND only); airport names are normalized to IATA codes before comparison (Appendix~\ref{sec:appendix_crossfamily}).}
\label{tab:cross_family_discordance}
\end{table}

\subsection{Prompt Perturbations are Bounded and Structured}
\label{sec:prompt_perturbations}

We next examine the sensitivity of LLM-generated distributions to controlled prompt perturbations. Table~\ref{tab:prompt_effects} reports JS divergence with bootstrap confidence intervals for different perturbation types. Persona conditioning induces bounded shifts\footnote{We use \emph{bounded} throughout in an empirical, comparative sense: these prompt-perturbation effects are smaller than the effect of changing model scale (Section~\ref{subsec:scale_discordance}), rather than as a claim of a provable theoretical bound on the divergence.} ($\mathrm{JS} = 0.13 \pm 0.01$), while demographic attributes produce larger effects, including age ($0.33 \pm 0.05$) and education ($0.26 \pm 0.03$). Context specification (regional vs.\ hub) also substantially reshapes distributions ($0.31 \pm 0.04$). The persona-conditioned distributions in Appendix~\ref{sec:appendix_personas} (Figure~\ref{fig:rq2_ind}) illustrate how persona information systematically redistributes probability mass across outcome types. Extended prompt-sensitivity analyses appear in Appendix~\ref{sec:appendix_sensitivity}.

A seasonality test (summer vs.\ winter) produces a smaller shift than any perturbation in Table~\ref{tab:prompt_effects} (mean JS divergence $= 0.098$, obtained by squaring each airport's distance in Table~\ref{tab:seasonality} and averaging), with large hubs showing smaller seasonal effects than regional airports, consistent with geographic anchoring (Appendix~\ref{sec:appendix_seasonality}).

\begin{table}[t]
\centering
\small
\setlength{\tabcolsep}{4pt}
\begin{tabular}{@{}lcc@{}}
\toprule
\textbf{Perturbation Type} & \textbf{JS Div.} & \textbf{N} \\
\midrule
Persona (with vs without) & $0.13 \pm 0.01$ & 298 \\
Age (within education)    & $0.33 \pm 0.05$ & 24 \\
Education (within age)    & $0.26 \pm 0.03$ & 60 \\
Context (regional vs hub) & $0.31 \pm 0.04$ & 40 \\
\bottomrule
\end{tabular}
\caption{Prompt sensitivity effect sizes (JS divergence; US airport travel demand). Mean $\pm$ half-CI over $N$ pairwise comparisons; see \S\ref{sec:bootstrap_protocol} for the precision convention.}
\label{tab:prompt_effects}
\end{table}

\subsection{Cross-domain generalization}
\label{sec:cross_domain_results}

To assess whether our findings generalize beyond the primary domain, we replicate our convergence analysis in two additional domains: restaurant choice (300 restaurants from the Yelp Open Dataset~\citep{yelp_dataset}, Indianapolis) and online product shopping (300 Home \& Kitchen products from Amazon metadata~\citep{ni2019justifying}, sampled via stratified embedding-based selection). Dataset construction is detailed in Appendix~\ref{sec:appendix_datasets}.

Across all three domains, the modes-first, tails-later pattern holds. The cumulative support keeps expanding as new tail items continue to appear, while the JS distance to the converged distribution decreases steadily, fully converging by iteration 8 for travel, 11 for Yelp, and 17 for Amazon. The three domains differ in single-context final support size (29 destinations for travel from Indianapolis vs.\ 103 restaurants and 51 products for Yelp and Amazon; across all 300 departure contexts the travel experiment reaches 185 unique destinations, Appendix~\ref{sec:appendix_results}) and in top-10 mass: Amazon (79\% of probability mass in the top-10) $>$ travel (75\%) $>$ Yelp (49\%), consistent with differences in utility structure.  The Amazon modes are a few specific best-selling appliances (e.g., a KitchenAid stand mixer, a Mr.\ Coffee espresso maker, and a Hamilton Beach slow cooker), whereas the Yelp mass spreads thinly across many interchangeable national chains (e.g., Starbucks, Chipotle, and Subway), indicating that product choices exhibit clearer utility hierarchies than restaurant preferences. Because Yelp and Amazon datasets are real-world choice sets whose specific item lists are unlikely to be memorized verbatim, and because the same dynamics appear across three model families trained on different corpora, this pattern is not attributable to single-corpus memorization.

\subsection{Robustness of Findings}
\label{sec:robustness}
\label{sec:temperature_sensitivity}
\label{sec:statistical_robustness}

We validate the central findings under four complementary robustness checks.

\paragraph{Temperature ablation.} Systematic ablation across 10 temperature values ($T \in \{0.1,\ldots,1.0\}$) and 6 settings confirms temperature effects are significant for maximum probability, entropy, and top-5 mass (6/6 settings, all $p<0.001$, Kruskal-Wallis) and directionally consistent in 17 of 18 metric-by-setting comparisons (Appendix~\ref{sec:appendix_temperature}). Low temperatures yield substantially greater ranking stability (Jaccard@5 = 0.93 at $T{=}0.2$ vs.\ 0.81 at $T{=}1.0$), validating $T{=}0.2$ as a principled choice; on those same three metrics the low-versus-high temperature contrast is large in 13 of 18 comparisons (Cohen's $|d| = 0.98$--$3.50$). Further, full greedy-decoding ($T{=}0$) replication preserves all main findings (Appendix~\ref{sec:appendix_greedy}).

\paragraph{Multi-source variability.} Across three orthogonal sources of variability, input-ordering permutations ($N{=}15$), prompt paraphrasing (5 variants), and multiple origin airports (IND, LAX, BOS), all metrics exhibit CV below 21\%, with the framework most robust to input ordering (mean CV = 9.6\%) and prompt paraphrasing (mean CV = 10.5\%). Core destinations emerge consistently regardless of ordering (e.g., ORD appears in top-5 across 100\% of permutations). Full per-source coefficients of variation (Table~\ref{tab:robustness_cv}) are reported in Appendix~\ref{sec:appendix_robustness}.

\paragraph{Variance decomposition.} An intraclass-correlation analysis (ICC(1,1); Appendix~\ref{sec:appendix_robustness}, Table~\ref{tab:icc_variance}) finds that departure airport and persona contribute little systematic between-condition variance (airport ICC $\le 0.08$ on every metric; persona $\le 0$ apart from a modest $0.16$ on peak probability), whereas model scale is the only factor with a non-trivial positive component, capturing the largest share of variance in distributional spread (support ICC $=0.46$) and the largest ICC on every metric except top-5 mass. This corroborates scale discordance through a variance-based lens, independent of the pairwise divergences in Section~\ref{subsec:scale_discordance}.

\section{Discussion and Limitations}
\label{sec:discussion}

Our results support viewing LLMs as structured stochastic generators of categorical distributions. The principal practical implication is that model choice, not prompt design, is the dominant lever shaping the elicited distribution: across three families and three domains, different models encode structurally different priors rather than refinements of one another, both between families and between scales within a family.

Several scope conditions limit these conclusions. The analysis is restricted to discrete choice settings with enumerable outcome spaces, specifically on the order of hundreds of outcomes, and the patterns may not transfer to open-ended generation. We evaluate open-weight models only, since closed-source APIs apply provider-side post-processing that confounds distributional measurement; whether the same dynamics hold in closed-source systems is an open question. All prompts are in English, and we do not benchmark against ground-truth human preferences, so the correspondence between characterized priors and real-world choice behavior is unstudied here. We emphasize, however, that in many settings where LLM-derived priors are most useful (small or hypothetical populations, future or counterfactual scenarios, and choices for which no survey or behavioral data exist), a ground-truth distribution is unobservable in principle rather than merely unavailable to us. Eliciting an LLM's implicit world prior is appealing precisely in such cases, where the operative question becomes the prior's internal consistency and structure (which we characterize) rather than its fidelity to a reference that does not exist. We also do not claim the observed regularities identify a specific internal mechanism: multiple internal processes may produce similar external behavior, and our contribution is to characterize that behavior systematically as consistent with implicit world priors. The margin by which model choice outweighs prompt design further depends on how far apart the two models are in scale: the divergence between the 8B model and either larger checkpoint ($0.474$ and $0.501$) separates from every prompt perturbation we measure with non-overlapping confidence intervals, whereas the 70B-versus-405B divergence ($0.365$) and the largest prompt effects (age, $0.330$; context, $0.310$) have intervals that overlap and are not resolved at our persona sample size. We read this as sharpening the claim rather than qualifying it: discordance scales with the size gap, so the practical guidance is strongest precisely where practitioners are most likely to substitute one model for another, and it identifies adjacent large checkpoints as the regime in which prompt design and model choice carry comparable weight.
\section{Conclusion and Future Work}

Treating LLMs as \emph{implicit world priors} (eliciting full probability distributions over discrete choice sets and analyzing them as repeated samples) reveals a consistent distributional anatomy across nine open-weight models, three families, and three choice preference domains. To this end, all models appear to be solid generators of distributions: distributions converge \emph{modes-first}, with top-$k$ outcomes stabilizing within a few samples while tail variability persists; for Llama, prompt perturbations induce systematic but bounded shifts below the effect of changing scale; and the modes-first dynamics replicate across all three choice domains and, in the travel domain, across all three model families. Decoding temperature amplifies or dampens this tail variability, but it does not create it: most of the tail is still present under greedy decoding ($T{=}0$), where sampling randomness is switched off entirely. On the other hand, there is a strong discordance between the outcome of the various models, both across families and scales.

Two practical implications follow. For downstream applications that treat LLM outputs as samples from a meaningful prior, model selection should be treated as a primary experimental variable, not as a cost--performance trade-off; and a small number of independent samples is sufficient for identifying dominant outcomes but not for characterizing tail uncertainty. Future work includes validating the elicited priors against real-world human preference data, developing theoretical accounts that link distributional dynamics to training objectives and data, and extending the framework to closed-source models, alternative architectures, and non-English settings. Until elicited priors can be checked against human preference data, we recommend reporting any LLM-derived distribution together with the model that produced it, since the distribution is as much a property of that model as of the question it was asked.

\section*{Ethical Considerations}

This work analyzes the distributional properties of LLM-generated outputs under controlled prompt variations, including demographic and contextual cues. While these analyses can reveal systematic patterns in model behavior, they are not intended to support normative or policy decisions. Any downstream use of LLM-generated distributions, particularly in decision-support or simulation settings, should be accompanied by validation against real-world data and appropriate domain expertise. Our findings further underline that model choice substantially influences the generated distribution, so transparency about model selection and sensitivity analyses is important whenever LLM outputs inform applied decisions. The experiments rely on synthetic data and do not involve personal information, but care should still be taken to avoid misinterpretation of demographically conditioned outputs. Large-scale LLM inference also incurs nontrivial computational and environmental costs, which we encourage practitioners to consider when planning experiments.

\section*{Acknowledgments}

We thank the developers of the open-weight LLMs evaluated here (Meta, Alibaba, Mistral AI) and the dataset providers (the Yelp Open Dataset and the Amazon Product Data collection released by the McAuley group), whose publicly released models and data made this study possible. AI writing assistants were used only for grammar and stylistic polish; all research design, analysis, and substantive writing are the authors' own. The experimental protocol was reviewed and determined to be exempt by the authors' institutional review board. Funding sources and IRB details will be released upon official acceptance.

\bibliography{custom}

\appendix

\section{Elicitation Method Validation: Full Comparison}
\label{sec:appendix_elicitation}

This section provides the full comparison of distribution elicitation methods summarized in Section~\ref{sec:elicitation_validation}.

\paragraph{Logit Inspection.}
Next-token logits provide probabilities over vocabulary tokens, not over multi-token outcomes such as three-letter IATA airport codes. Constructing a joint distribution over 300 airports from logits would require evaluating conditional token chains for each candidate (300--900 forward passes per distribution), and most API providers expose only logarithms of probabilities of top-$N$ results rather than the full vocabulary distribution.

\paragraph{Repeated Single-Choice Sampling.}
Building a distribution by sampling individual choices (e.g., ``name one destination'') would require thousands of API calls per distribution to reliably estimate tail probabilities. More fundamentally, pilot experiments reveal that this approach fails to capture distributional structure. At $T{=}0.2$, 300 independent single-choice calls returned the same destination (ORD) in 100\% of cases, yielding a delta function with zero entropy. Even at $T{=}1.0$ (maximum temperature), only 2 unique destinations appeared across 300 calls (ORD 97\%, MDW 3\%), yielding entropy of 0.18 bits. In contrast, verbalized distributions from a single call at the same temperatures captured 34--35 unique destinations with entropy of 3.97--4.25 bits. Table~\ref{tab:elicitation_comparison} summarizes the comparison.

\begin{table}[t]
\centering
\small
\setlength{\tabcolsep}{2pt}
\begin{tabular}{@{}llcccc@{}}
\toprule
\textbf{Method} & \textbf{T} & \emph{Sup.} & \emph{Entropy} & \emph{Top-5} & \emph{JS} \\
\midrule
Single-choice & 0.2 & 1 & $0.00{\pm}0.00$ & $1.00{\pm}0.00$ & $0.64{\pm}0.02$ \\
Single-choice & 1.0 & 2 & $0.18{\pm}0.10$ & $1.00{\pm}0.00$ & $0.61{\pm}0.04$ \\
Verbalized & 0.2 & 35 & $3.97{\pm}0.29$ & $0.58{\pm}0.06$ & (ref.) \\
Verbalized & 1.0 & 34 & $4.25{\pm}0.20$ & $0.46{\pm}0.04$ & (ref.) \\
\bottomrule
\end{tabular}
\caption{Comparison of distribution elicitation methods for IND departures (Llama-3.3-70B). \emph{Sup.}: unique destinations observed. \emph{Entropy}: Shannon entropy in bits. \emph{Top-5}: probability mass in top 5 destinations. \emph{JS}: Jensen--Shannon divergence against the verbalized baseline at the same temperature; ``(ref.)'' marks the verbalized rows, which \emph{are} that baseline (self-divergence $0$). Entropy, Top-5 and JS are mean $\pm$ half-CI over $10{,}000$ bootstrap resamples of the underlying calls (20 verbalized calls at $T{=}0.2$ and 18 at $T{=}1.0$; 300 single-choice draws at each). Support is an exact count over the full sample and is given without an interval, because resampling calls can only remove distinct destinations and never add them, so its bootstrap is biased downward. The zero intervals on the single-choice rows are structural rather than rounding: at $T{=}0.2$ every draw returned the same destination, and with a support of one or two all mass necessarily falls inside the top five. Single-choice sampling produces degenerate distributions at both temperatures, while verbalized elicitation captures rich distributional structure in a single query.}
\label{tab:elicitation_comparison}
\end{table}

The JS divergence between single-choice and verbalized distributions is 0.64 at $T{=}0.2$ and 0.61 at $T{=}1.0$, confirming that the two methods capture fundamentally different information: single-choice reflects the model's point estimate (which single destination it would pick), while verbalized distributions express the model's probability beliefs over the entire outcome space. Verbalized elicitation is also far more sample-efficient: a single call yields a support of 35 destinations, whereas the 300 single-choice calls yield 2 (Table~\ref{tab:elicitation_comparison}).

\section{Detailed Metric Definitions}
\label{sec:appendix_metrics}

This section provides complete mathematical definitions for the evaluation metrics summarized in Section~\ref{sec:evaluation_metrics}. All pairwise metrics (JS divergence, Jaccard, Spearman) are computed between pairs of independent samples $p^{(r)}$ and $p^{(r')}$; since samples are independent, the ordering of $r$ and $r'$ is arbitrary. When reported in tables, values are averaged across all pairs within a condition and across departure airports unless otherwise noted.

\subsection{Distributional Consistency Metrics}

\paragraph{Jensen--Shannon Divergence.} To measure overall distributional similarity between runs, we use Jensen--Shannon (JS) divergence \citep{lin1991divergence}:
\begin{equation}
\begin{split}
\mathrm{JS}(p^{(r)}, p^{(r')}) = {} & \frac{1}{2} D_{\mathrm{KL}}(p^{(r)} \| m) \\
& + \frac{1}{2} D_{\mathrm{KL}}(p^{(r')} \| m)
\end{split}
\end{equation}
where $m = \frac{1}{2}(p^{(r)} + p^{(r')})$ is the mixture distribution and $D_{\mathrm{KL}}$ denotes Kullback--Leibler divergence. JS divergence is symmetric and well-suited for comparing discrete distributions. Logarithms are natural except where noted. Tables~\ref{tab:mode_tail_decomp}, \ref{tab:model_agreement}, \ref{tab:cross_family_discordance}, \ref{tab:prompt_effects} and~\ref{tab:k_sensitivity} report this divergence, bounded in $[0, \ln 2]$. Table~\ref{tab:elicitation_comparison} reports it with the base-2 logarithm (bound $[0,1]$), matching its entropy column in bits. Tables~\ref{tab:cross_domain}, \ref{tab:seasonality} and~\ref{tab:greedy_comparison}, and the JS panel of Figure~\ref{fig:cross_domain}, report the Jensen--Shannon \emph{distance}, the square root of the divergence, bounded by $\sqrt{\ln 2}$. Distances and divergences are not directly comparable, so any comparison across these groups states the conversion it uses. Lower values indicate greater similarity; JS $= 0$ implies identical distributions.

\paragraph{Mode Stability (Top-$k$ Invariance).} We measure mode stability using the Jaccard index \citep{jaccard1912distribution}:
\begin{equation}
\mathrm{Jaccard}@k(r,r') = \frac{|\mathcal{M}_k(p^{(r)}) \cap \mathcal{M}_k(p^{(r')})|}{|\mathcal{M}_k(p^{(r)}) \cup \mathcal{M}_k(p^{(r')})|}
\end{equation}
where $r$ and $r'$ denote two sampling runs, $\mathcal{M}_k(p^{(r)})$ is the set of top-$k$ outcomes from run $r$, $\cap$ and $\cup$ denote set intersection and union, and $|\cdot|$ denotes set cardinality. Jaccard ranges from 0 (no overlap) to 1 (identical sets). Higher values indicate stable dominant outcomes across samples.

\paragraph{Rank Consistency.} For ranked mode sets, we compute Spearman rank correlation \citep{spearman1904proof} over the top-$k$ outcomes to assess ordering stability. Spearman's $\rho$ ranges from $-1$ to $+1$, where $+1$ indicates perfect rank agreement and values near 0 indicate no systematic relationship.

\subsection{Tail Dynamics Metrics}

We employ standard information-theoretic measures \citep{cover2006elements} to characterize tail behavior:

\paragraph{Support Growth.}
\begin{equation}
|\{ y : p^{(r)}(y) > \epsilon \}|
\end{equation}
This counts the number of outcomes with non-negligible probability in a single sample (we use $\epsilon = 0.001$). Higher values indicate broader distributional coverage. In tables, support is reported per sample; the growth percentage compares the first sample's support to the cumulative support after all samples (Unique@$R$).

\paragraph{Entropy.}
\begin{equation}
H(p^{(r)}) = -\sum_{y} p^{(r)}(y) \log p^{(r)}(y)
\end{equation}
Shannon entropy measures distributional uncertainty of a single sample \citep{cover2006elements}. Higher entropy indicates more uniform (less concentrated) distributions; entropy is minimized at 0 for deterministic distributions. In tables, entropy values are reported per sample or averaged across samples within a condition.

\paragraph{Tail Mass.}
\begin{equation}
\mathrm{TailMass}_k(p^{(r)}) = 1 - \sum_{y \in \mathcal{M}_k} p^{(r)}(y)
\end{equation}
Tail mass is defined as the complement of the cumulative probability in the mode set $\mathcal{M}_k$, following standard probability concentration analysis \citep{cover2006elements}. It quantifies the probability allocated to non-mode outcomes. Values range from 0 (all mass in top-$k$) to near 1 (highly dispersed). Higher values indicate greater exploration of low-probability alternatives.

\subsection{Mode--Tail Divergence Analysis}

To localize sources of variability, we compute JS divergence \citep{lin1991divergence} separately on the mode and tail regions of the distribution. Given any two independent samples $p^{(r)}$ and $p^{(r')}$, we define three metrics:

\emph{Total JS Divergence} measures overall distributional shift:
\begin{equation}
\mathrm{JS}_{\text{total}} = \mathrm{JS}(p^{(r)}, p^{(r')})
\end{equation}

\emph{Mode JS Divergence} measures variability in the high-probability region. We restrict both distributions to the union of their top-$k$ sets $\mathcal{M}_k(p^{(r)}) \cup \mathcal{M}_k(p^{(r')})$, renormalize each to sum to 1, and compute:
\begin{equation}
\mathrm{JS}_{\text{mode}} = \mathrm{JS}(\tilde{p}^{(r)}_{\mathcal{M}}, \tilde{p}^{(r')}_{\mathcal{M}})
\end{equation}
where $\tilde{p}_{\mathcal{M}}$ denotes the renormalized distribution over $\mathcal{M}_k$.

\emph{Tail JS Divergence} measures instability in the low-probability region, computed analogously on $\mathcal{T}_k = \mathcal{Y} \setminus \mathcal{M}_k$:
\begin{equation}
\mathrm{JS}_{\text{tail}} = \mathrm{JS}(\tilde{p}^{(r)}_{\mathcal{T}}, \tilde{p}^{(r')}_{\mathcal{T}})
\end{equation}

\emph{Important:} These three metrics are computed independently on different subsets of the outcome space. They do not form an additive decomposition (i.e., $\mathrm{JS}_{\text{total}} \neq \mathrm{JS}_{\text{mode}} + \mathrm{JS}_{\text{tail}}$). Rather, they provide complementary views: $\mathrm{JS}_{\text{total}}$ captures overall divergence, while $\mathrm{JS}_{\text{mode}}$ and $\mathrm{JS}_{\text{tail}}$ isolate divergence within each region after renormalization. This allows us to identify whether instability arises from semantic core beliefs (mode divergence) or from peripheral exploration (tail divergence).

\emph{Interpretation of metric magnitudes:} It is expected that $\mathrm{JS}_{\text{tail}} > \mathrm{JS}_{\text{total}}$ in practice. The tail region contains many low-probability outcomes that are inherently noisy; after renormalization (scaling tail probabilities to sum to 1), small absolute differences become large relative differences, magnifying the measured divergence. Conversely, $\mathrm{JS}_{\text{mode}} < \mathrm{JS}_{\text{total}}$ because high-probability outcomes stabilize quickly and exhibit less variability. The key diagnostic insight is the \emph{trajectory} of these metrics: when $\mathrm{JS}_{\text{mode}}$ approaches zero while $\mathrm{JS}_{\text{tail}}$ remains elevated, this indicates that core semantic beliefs have converged and only peripheral outcomes continue to fluctuate.

\subsection{Prompt Sensitivity Analysis}

To quantify distributional shift under controlled prompt perturbations, we compute the JS divergence \citep{lin1991divergence} between the two conditions \emph{for each comparison unit} and report the mean over units, $\frac{1}{|U|}\sum_{u \in U} \mathrm{JS}\!\left(p_A^{(u)}, p_B^{(u)}\right)$, where $p_A^{(u)}$ and $p_B^{(u)}$ are the normalized distributions under conditions $A$ and $B$ for unit $u$, and $U$ is the set of comparison units; $95\%$ confidence intervals are obtained by bootstrapping over units. The unit depends on the perturbation: a departure context for persona vs.\ no-persona; a pair of personas differing in one demographic attribute, with the other held fixed, for age and education; and a persona evaluated at a regional versus a hub airport for location. Higher JS values indicate greater sensitivity to the perturbation, allowing us to quantify and compare prompt effect sizes.

\begin{table*}[h]
\centering
\small
\begin{tabular}{@{}llcc@{}}
\toprule
\textbf{Family} & \textbf{Model ID} & \textbf{Params} & \textbf{Role} \\
\midrule
\multirow{3}{*}{Llama} & Llama-3.1-8B-Instruct-Turbo & 8B & Smaller \\
 & Llama-3.3-70B-Instruct-Turbo & 70B & Baseline \\
 & Meta-Llama-3.1-405B-Instruct-Turbo & 405B & Larger \\
\midrule
\multirow{3}{*}{Qwen} & Qwen2.5-7B-Instruct-Turbo & 7B & Small \\
 & Qwen2.5-72B-Instruct-Turbo & 72B & Medium \\
 & Qwen3-235B-A22B-Instruct & 235B (MoE) & Large \\
\midrule
\multirow{3}{*}{Mistral} & Mistral-7B-Instruct-v0.3 & 7B & Small \\
 & Mistral-Small-24B-Instruct & 24B & Medium \\
 & Mixtral-8x7B-Instruct-v0.1 & 46B (MoE) & Large \\
\bottomrule
\end{tabular}
\caption{Complete model configuration across three families. All models accessed via the Together AI API with temperature $T=0.2$.}
\label{tab:model_config}
\end{table*}

\subsection{Scale Discordance Analysis}

To assess whether different model scales encode consistent world priors, we compare distributions across models $m_i, m_j$ using JS divergence \citep{lin1991divergence}:
\begin{equation}
\mathrm{JS}(p_{m_i}(\cdot \mid x), p_{m_j}(\cdot \mid x))
\end{equation}

We additionally compute agreement@$k$, defined as the proportion of contexts where two models share identical top-$k$ outcome sets. Agreement@$k$ ranges from 0 (no shared predictions) to 1 (perfect agreement); values close to the chance level (which is small relative to 1 when the outcome space is large) indicate independent distributional priors.

\subsection{Justification for $k{=}5$ Mode--Tail Threshold}
\label{sec:appendix_k_justification}

We adopt a fixed-$k$ definition rather than a cumulative-probability threshold (top-$p$) because fixed set sizes are essential for direct cross-model and cross-condition comparisons via Jaccard overlap and rank correlation, the central metrics in our scale discordance and prompt sensitivity analyses. A top-$p$ criterion would produce variable-size mode sets (e.g., 3 destinations for a peaked distribution vs.\ 50 for a flat one), confounding set-size differences with genuine distributional disagreement.

The choice of $k{=}5$ is empirically grounded. We identify the natural mode--tail boundary by computing the probability gap between each consecutively ranked outcome (i.e., the difference $p(y_{(i)}) - p(y_{(i+1)})$ where outcomes are sorted by decreasing probability); the rank at which this gap is largest indicates where the distribution transitions from concentrated modes to a diffuse tail. Across 300 departure contexts, this largest gap falls at rank $\leq 5$ in 99\% of distributions (median at rank~1).

Table~\ref{tab:k_sensitivity} repeats the mode--tail analysis of Table~\ref{tab:mode_tail_decomp} for $k \in \{3, 5, 7, 10, 15\}$ across the four primary contexts. Two of the three $k$-dependent quantities are insensitive to the cutoff: tail divergence exceeds mode divergence in both post-convergence phases at every $k \geq 5$, with the margin widening as $k$ grows (in the 10--20 phase, from $2.3\times$ at $k{=}5$ to $3.7\times$ at $k{=}15$) because outcomes beyond the natural boundary are then counted as modes; at $k{=}3$ the gap survives only in the later phase, as expected when the cutoff sits below the natural elbow; and post-convergence mode agreement is nearly flat (mean Jaccard@$k$ between $0.72$ and $0.77$). The stabilization point $R^*$ is the exception, drifting later as $k$ grows ($1$--$7$ at $k{=}3$ versus $10$--$18$ at $k{=}15$). This drift is mechanical rather than substantive: a single swap in a top-$k$ set pushes the Jaccard index below the detection threshold for any $k$ (Appendix~\ref{sec:appendix_stab_threshold}), and larger mode sets offer more boundary outcomes to swap. It is a further reason to fix $k$ at the natural elbow rather than above it.

The sweep above is computed at the primary setting $T{=}0.2$, but the mode--tail structure is not an artifact of that choice. Across the ten-value temperature ablation of Appendix~\ref{sec:appendix_temperature}, tail divergence exceeds mode divergence at every temperature tested ($10$ of $10$ values in $T \in [0.1, 1.0]$), and mode-set stability is preserved at both cutoffs: Jaccard@10 stays between $0.81$ and $0.91$ while Jaccard@5 decays from $0.93$ to $0.81$, and top-5 mass remains between $0.49$ and $0.57$ (Table~\ref{tab:temperature_summary}). Raising the temperature therefore blurs the mode--tail separation gradually rather than relocating the boundary.

\begin{table*}[t]
\centering
\small
\setlength{\tabcolsep}{3pt}
\renewcommand{\arraystretch}{0.95}
\begin{tabular}{@{}lccccccc@{}}
\toprule
 & \multicolumn{2}{c}{Iter 5--10} & \multicolumn{2}{c}{Iter 10--20} & & & \\
\cmidrule(lr){2-3}\cmidrule(lr){4-5}
$k$ & $\mathrm{JS}_{\mathrm{m}}$ & $\mathrm{JS}_{\mathrm{t}}$ & $\mathrm{JS}_{\mathrm{m}}$ & $\mathrm{JS}_{\mathrm{t}}$ & Jac@$k$ & TM@$k$ & $R^*$ \\
\midrule
3  & $0.07{\pm}0.04$ & $0.09{\pm}0.03$ & $0.06{\pm}0.03$ & $0.10{\pm}0.03$ & 0.77 & 54.4--57.6 & 1--7 \\
5  & $0.06{\pm}0.04$ & $0.12{\pm}0.04$ & $0.06{\pm}0.03$ & $0.13{\pm}0.04$ & 0.77 & 36.7--40.2 & 1--14 \\
7  & $0.06{\pm}0.03$ & $0.18{\pm}0.05$ & $0.06{\pm}0.03$ & $0.18{\pm}0.05$ & 0.75 & 23.9--27.2 & 1--20 \\
10 & $0.06{\pm}0.03$ & $0.27{\pm}0.09$ & $0.07{\pm}0.03$ & $0.24{\pm}0.06$ & 0.76 & 11.9--14.9 & 3--17 \\
15 & $0.07{\pm}0.03$ & $0.23{\pm}0.12$ & $0.08{\pm}0.03$ & $0.29{\pm}0.09$ & 0.72 & 3.9--5.8 & 10--18 \\
\bottomrule
\end{tabular}
\caption{Sensitivity of the mode--tail analysis to the cutoff $k$ (US airport travel demand; four primary departure contexts). $\mathrm{JS}_{\mathrm{m}}$ and $\mathrm{JS}_{\mathrm{t}}$ are the mode- and tail-region divergences by convergence phase, computed exactly as in Table~\ref{tab:mode_tail_decomp}; the $k{=}5$ row reproduces that table. Divergences are mean $\pm$ half-CI over the consecutive-sample comparisons pooled across the four contexts (\S\ref{sec:bootstrap_protocol}). Jac@$k$: mean top-$k$ Jaccard between consecutive post-convergence samples. TM@$k$: percentage of probability mass outside the top-$k$ modes. $R^*$: cumulative top-$k$ stabilization point at $\tau{=}0.9$. TM@$k$ and $R^*$ are given as ranges over the four contexts. Tail dominance holds for $k \geq 5$, where the paired per-comparison difference $\mathrm{JS}_{\mathrm{t}} - \mathrm{JS}_{\mathrm{m}}$ excludes zero in both post-convergence phases; at $k{=}3$ it holds for iterations 10--20 ($+0.039$, CI $[+0.008, 0.068]$) but not for 5--10 ($+0.024$, CI $[-0.014, 0.058]$), which is expected because $k{=}3$ truncates the mode region below its natural elbow. Mode agreement is flat in $k$, whereas $R^*$ drifts later with $k$ for the mechanical reason given in the text.}
\label{tab:k_sensitivity}
\end{table*}

\subsection{Threshold-Insensitivity of the Stabilization Point}
\label{sec:appendix_stab_threshold}

The stabilization point $R^*$ (Section~\ref{sec:evaluation_metrics}) is detected by thresholding the top-$k$ Jaccard index between successive cumulative top-$k$ sets, and the specific value of this threshold is immaterial. For $k{=}5$, the Jaccard index of two size-$5$ sets can only take the values $\{0,\,0.11,\,0.25,\,0.43,\,0.67,\,1.0\}$: it equals $1$ when the sets are identical and drops to at most $0.67$ as soon as a single element is swapped, leaving the band $(0.67,\,1.0)$ unattainable. Consequently, the recovered stabilization point is constant for \emph{every} threshold $\tau$ in this band: across all four primary contexts, $R^*(\tau)$ is flat throughout $(0.67,\,1.0)$, so the choice of $0.9$ has no effect on the result and is not a tunable parameter.

\section{Experimental Configuration Details}
\label{sec:appendix_experimental}

\begin{figure*}[t]
\centering
\includegraphics[width=\textwidth]{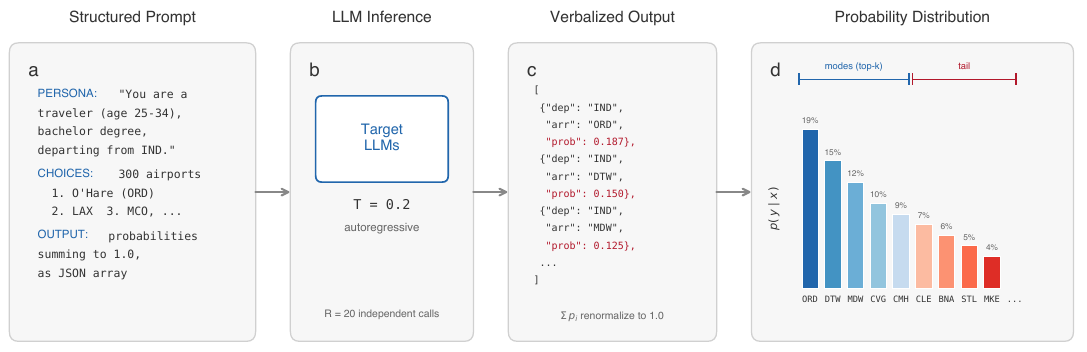}
\caption{Overview of the experimental pipeline. (a)~A structured prompt provides the LLM with a traveler persona (age, education), a departure airport, and 300 candidate destination airports, requesting probability assignments as a JSON array. (b)~The target LLM generates a response autoregressively at temperature $T{=}0.2$; this process is repeated $R{=}20$ times as independent API calls. (c)~Each raw JSON output is parsed and probabilities are renormalized to sum to 1.0. (d)~The resulting probability distribution over destinations, shown here for Indianapolis (IND) using real experimental data. High-probability destinations (blue, \emph{modes}) and low-probability destinations (red, \emph{tail}) are analyzed for distributional consistency, mode--tail dynamics, scale discordance, and prompt sensitivity. See Appendix~\ref{sec:appendix_prompts} for complete prompt templates.}
\label{fig:pipeline}
\end{figure*}

This section provides complete details for the experimental setup summarized in Section~\ref{sec:experimental_setup}.

\subsection{Persona Configuration}

We define personas using the following age and education categories. The American Community Survey~\citep{acs2025} provides demographic background; the categories are specified directly in the experiment code:

\emph{Age Groups (5 levels):}
\begin{itemize}
    \item 18--24 (Young)
    \item 25--34 (Young Adult)
    \item 35--44 (Adult)
    \item 45--64 (Middle Age)
    \item 65 and above (Senior)
\end{itemize}

\emph{Education Levels (4 levels):}
\begin{itemize}
    \item Under High School
    \item High School to Under Bachelor
    \item Bachelor Degree
    \item Above Bachelor (Master/Doctoral)
\end{itemize}

This yields 20 unique persona combinations ($5 \times 4 = 20$). All combinations are included as prompt conditions. No ACS population estimates are used to construct or weight the grid, and no specific ACS table or release vintage is attributed to it.

\subsection{Prompt Templates}
\label{sec:appendix_prompts}

Each LLM query is constructed from four components: a \emph{persona}, a \emph{choice set}, a \emph{question}, and an \emph{instruction} specifying the output format. Below we reproduce the templates for each experimental condition, with uppercase section labels added here for readability and long choice sets elided; placeholder fields (e.g., \texttt{\{age\_range\}}) are filled dynamically.

\paragraph{Base Prompt (Travel Demand, Persona-Conditioned).}
\begin{small}
\begin{verbatim}
PERSONA: You are a traveler (age {age_range})
  with {education_desc}, departing from
  {departure_state} ({departure_name}).

AVAILABLE AIRPORTS:
1. {airport_name} ({code}) - {state}
2. {airport_name} ({code}) - {state}
... [300 airports listed]

QUESTION: Choose your travel destinations from
  the airports listed above. What are your
  travel plans and their probabilities?

INSTRUCTION: Please plan travel for the next
  12 months based on your demographic profile.
  Provide probabilities as float values from
  0 to 1, where higher values indicate higher
  likelihood. All travel probabilities should
  sum to 1.0. Focus only on destination choices
  regardless of connection routes.
  Format your response as:
  [{"departure_airport":, "arrival_airport":,
    "travel_probability":}, ...]
\end{verbatim}
\end{small}

\paragraph{No-Persona Variant (Control).}
The persona field is reduced to: \texttt{``You are a traveler departing from \{departure\_state\} (\{departure\_name\}).''} All other components remain identical.

\paragraph{Seasonal Variant.}
The persona is appended with: \texttt{``You are planning travel for \{season\_context\}.''} where \texttt{season\_context} is either ``summer (June--August)'' or ``winter (December--February).'' The question is modified to: ``What are your travel plans and their probabilities for this season?''

\paragraph{Restaurant Choice (Yelp).}
\begin{small}
\begin{verbatim}
PERSONA: You are a diner (age {age_range})
  with {education_desc}, living in {city}.

AVAILABLE RESTAURANTS:
1. {restaurant_name} ({cuisine_category})
... [300 restaurants listed]

QUESTION: Choose your dining preferences from
  the restaurants listed above. What is your
  probability of dining at each restaurant
  in the next month?

INSTRUCTION: Please indicate your restaurant
  preferences based on your demographic profile.
  Provide probabilities as float values from 0
  to 1, where higher values indicate higher
  likelihood of dining there. All probabilities
  should sum to 1.0. Format your response as a
  JSON array (no other text before or after):
  [{"restaurant": "Restaurant Name",
    "probability": 0.05}, ...]
\end{verbatim}
\end{small}

\paragraph{Product Choice (Amazon).}
\begin{small}
\begin{verbatim}
PERSONA: You are an online shopper
  (age {age_range}) with {education_desc}.

AVAILABLE PRODUCTS:
1. {product_name} ({product_category})
... [300 products listed]

QUESTION: Choose your product preferences from
  the items listed above. What is your
  probability of purchasing each product
  in the next month?

INSTRUCTION: Please indicate your product
  preferences based on your demographic profile.
  Provide probabilities as float values from 0
  to 1, where higher values indicate higher
  likelihood of purchasing. All probabilities
  should sum to 1.0. Format your response as a
  JSON array (no other text before or after):
  [{"product": "Product Name",
    "probability": 0.05}, ...]
\end{verbatim}
\end{small}

\paragraph{Expected Output Format.}
The LLM generates a JSON array as free-form text. For the travel domain, a typical output looks like:
\begin{small}
\begin{verbatim}
[
  {"departure_airport": "IND",
   "arrival_airport": "ORD",
   "travel_probability": 0.15},
  {"departure_airport": "IND",
   "arrival_airport": "LAX",
   "travel_probability": 0.12},
  {"departure_airport": "IND",
   "arrival_airport": "MCO",
   "travel_probability": 0.08},
  ...
]
\end{verbatim}
\end{small}
Raw probability sums before renormalization are frequently not equal to $1.0$, and the deviation is model-dependent. Across the nine models (Indianapolis departures, $20$ samples per model; $180$ outputs), the raw sums have a median of $0.98$, but only $40\%$ fall within $[0.95, 1.05]$ ($52\%$ within $[0.9, 1.1]$); per-model medians range from $0.81$ (Llama-3.3-70B) to $1.29$ (Llama-3.1-8B), with some models (e.g., Qwen3-235B) consistently near $1.0$ while smaller models tend to over- or under-shoot. We therefore renormalize every parsed output to sum to exactly $1.0$. See Appendix~\ref{sec:appendix_parsing} for the complete parsing and rectification pipeline.

\subsection{Research Question Details}

\emph{Distributional Consistency and Convergence:} We perform 20 iterations at multiple departure contexts (IND, FWA, ATL, ORD) using a fixed persona, measuring JS divergence (overall and within the mode and tail regions) and Jaccard@$k$ stability. Extension experiments test convergence at hub contexts and across model sizes.

\emph{Sensitivity to Prompt Perturbations:} We test perturbations including: (a) persona demographics (age, education), (b) temporal context (season), and (c) geographic anchors (regional vs hub). We compute JS divergence with bootstrap confidence intervals. Seasonality experiments compare summer (June--August) vs winter (December--February) across 4 airports with 20 iterations per condition (160 total runs).

\emph{Scale Discordance and Generalization:} We compare distributions across model sizes using pairwise JS divergence and agreement@$k$. Cross-family analysis extends to the Qwen and Mistral families to test universality. Cross-domain replication uses restaurant choice (Yelp, 300 Indianapolis restaurants) and product shopping (Amazon, 300 Home \& Kitchen products). See Appendix~\ref{sec:appendix_datasets} for dataset details.

\section{Cross-Domain Dataset Descriptions}
\label{sec:appendix_datasets}

This section provides detailed descriptions of the datasets and sampling procedures used for the cross-domain generalization experiments.

\subsection{Yelp Restaurant Dataset}

\emph{Source:} Yelp Open Dataset~\citep{yelp_dataset} (\url{https://business.yelp.com/data/resources/open-dataset/})

\emph{Data Extraction:} We extracted restaurant data from the Yelp Open Dataset, focusing on Indianapolis, Indiana, to match the geographic context of our primary travel demand experiments.

\emph{Filtering Criteria:}
\begin{itemize}
    \item \emph{Location:} Businesses located in Indianapolis (city = ``Indianapolis'')
    \item \emph{Category:} Businesses whose Yelp category list matches any of 22 dining keywords (for example \emph{restaurant}, \emph{cafe}, \emph{pizza}, \emph{seafood}), excluding seven non-restaurant food keywords (for example \emph{grocery}, \emph{catering}, \emph{food truck})
    \item \emph{Status:} Only open businesses (is\_open = 1)
    \item \emph{Name Quality:} Excluded entries with generic or missing names
    \item \emph{Category Quality:} Excluded entries with only generic categories (e.g., ``Restaurants'' alone without cuisine specification)
\end{itemize}

\emph{Final Dataset:}
\begin{itemize}
    \item 300 restaurants sampled from Indianapolis
    \item 54 unique cuisine categories represented
    \item Each restaurant record includes: name, cuisine type, star rating, review count, and location attributes
\end{itemize}

\emph{Cuisine Distribution:} The extracted restaurants span diverse cuisine types including American (Traditional), American (New), Mexican, Italian, Chinese, Japanese, Indian, Thai, Mediterranean, and 45 other cuisine categories, providing a semantically rich choice set for LLM evaluation.

\subsection{Amazon Product Dataset}

\emph{Source:} Amazon Product Data~\citep{ni2019justifying} (\url{https://nijianmo.github.io/amazon/index.html})

\emph{Category:} Home \& Kitchen product metadata

\emph{Sampling Procedure:} To ensure semantic diversity while maintaining manageable choice set size, we implemented a stratified embedding-based sampling approach:

\begin{enumerate}
    \item \emph{Category Stratification:} Products were grouped by subcategory within Home \& Kitchen, yielding 15 subcategories (e.g., Cookware \& Bakeware, Bedding, Bath, Storage \& Organization, Home Decor, etc.)

    \item \emph{Embedding Generation:} Product titles concatenated with brand names were embedded using OpenAI's \texttt{text-embedding-3-large} model to capture semantic similarity

    \item \emph{Clustering:} Within each subcategory, K-means clustering was applied to identify semantically distinct product groups

    \item \emph{Representative Sampling:} Products closest to each cluster centroid were selected as representatives, ensuring coverage of diverse product types within each subcategory

    \item \emph{Near-Duplicate Removal:} Products with embedding cosine similarity $> 0.90$ were filtered to prevent redundant entries

    \item \emph{Balanced Selection:} 20 products were sampled from each of the 15 subcategories, yielding 300 total products
\end{enumerate}

\emph{Final Dataset:}
\begin{itemize}
    \item 300 Home \& Kitchen products
    \item 15 subcategories with balanced representation (20 products each)
    \item Each product record includes: title, description, category hierarchy, price, and rating attributes
\end{itemize}

\subsection{Domain Comparison}

The three domains used in this study differ in key dimensions (Table~\ref{tab:domain_comparison}):

\begin{table}[h]
\centering
\small
\begin{tabular}{@{}ll@{}}
\toprule
\textbf{Characteristic} & \textbf{Value} \\
\midrule
\multicolumn{2}{c}{\textit{Travel Domain}} \\
\midrule
Choice Set Size & 300 airports \\
Geographic Scope & National (US) \\
Decision Frequency & Low (occasional) \\
Price Range & High (\$100 and above) \\
Category Type & Transportation \\
\midrule
\multicolumn{2}{c}{\textit{Yelp Domain}} \\
\midrule
Choice Set Size & 300 restaurants \\
Geographic Scope & Local (Indianapolis) \\
Decision Frequency & Medium (weekly) \\
Price Range & Medium (\$10-50) \\
Category Type & Food \& Dining \\
\midrule
\multicolumn{2}{c}{\textit{Amazon Domain}} \\
\midrule
Choice Set Size & 300 products \\
Geographic Scope & N/A (online) \\
Decision Frequency & High (frequent) \\
Price Range & Variable \\
Category Type & Consumer Goods \\
\bottomrule
\end{tabular}
\caption{Comparison of domain characteristics across experimental settings.}
\label{tab:domain_comparison}
\end{table}

This diversity enables testing whether the observed distributional patterns (modes-first convergence, scale discordance, prompt sensitivity bounds) generalize beyond the primary travel demand setting.

\subsection{Cross-Domain Convergence Comparison}

Table~\ref{tab:cross_domain} consolidates per-metric convergence numbers across the three domains. All values come from single-context, 20-iteration convergence runs: travel from the Indianapolis context (Table~\ref{tab:convergence}), and Yelp and Amazon from their respective single-context experiments. All domains provide 300 options; \emph{final support} counts unique items with probability $>$0.1\% in the converged distribution (for travel, this $>$0.1\% count, 29, is slightly tighter than the non-zero Unique@20 of $32$ in Table~\ref{tab:convergence}).

\begin{table}[h]
\centering
\small
\setlength{\tabcolsep}{3pt}
\begin{tabular}{@{}lccc@{}}
\toprule
\textbf{Metric} & \textbf{Travel} & \textbf{Yelp} & \textbf{Amazon} \\
\midrule
Options & 300 & 300 & 300 \\
JS converge (iter) & 8 & 11 & 17 \\
Final support & 29 & 103 & 51 \\
Top-10 mass (\%) & 74.7\% & 49.4\% & 79.0\% \\
\bottomrule
\end{tabular}
\caption{Cross-domain comparison of convergence metrics across travel demand, Yelp restaurants, and Amazon products. All three are single-context, $20$-iteration convergence runs: travel uses the Indianapolis (IND) context (Table~\ref{tab:convergence}), matching the single-context Yelp and Amazon experiments; \emph{JS converge} is the first iteration at which the JS distance to the converged distribution falls below $0.05$, applied identically across domains. \emph{Top-10 mass} is the probability mass carried by the ten highest-probability items of the converged, renormalized distribution; note the cutoff here is $k{=}10$ rather than the $k{=}5$ mode boundary used elsewhere (Section~\ref{subsec:mode_tail}), so that all three domains are reported at a single common cutoff. The broader travel result across all $300$ departure contexts ($185$ unique destinations) is reported separately in Section~\ref{sec:appendix_results}. Referenced from \S\ref{sec:cross_domain_results} in the main text.}
\label{tab:cross_domain}
\end{table}

\section{Parsing Methodology for Smaller Models}
\label{sec:appendix_parsing}

Smaller LLMs (8B parameters) exhibit different output formatting behaviors compared to larger models, requiring specialized parsing strategies. This section documents the parsing challenges encountered and the rectification methodology applied, and reports the unrectified baseline for transparency (\S\ref{sec:appendix_original_results}).

\subsection{Original Parsing Failures}

With the original single-tier parsing approach (JSON bracket extraction only), the smaller model (Llama-3.1-8B) exhibited significant parsing failures (Table~\ref{tab:parsing_comparison}):

\begin{table}[h]
\centering
\small
\begin{tabular}{@{}lcc@{}}
\toprule
\textbf{Metric} & \textbf{Original} & \textbf{Rectified} \\
\midrule
\multicolumn{3}{c}{\textit{Persona-based Analysis}} \\
\midrule
Success Rate & 15\% (3/20) & 100\% (20/20) \\
Unique Destinations & 27 & 131 \\
Avg Destinations/Persona & 13.0 & 23.9 \\
Total Records & 39 & 477 \\
\midrule
\multicolumn{3}{c}{\textit{Convergence Analysis}} \\
\midrule
Iterations Parsed & 6/20 & 20/20 \\
Total Destinations & 63 & 308 \\
Final Unique Destinations & --- & 100 \\
\bottomrule
\end{tabular}
\caption{Comparison of smaller model (8B) results before and after parsing rectification.}
\label{tab:parsing_comparison}
\end{table}

\subsection{Failure Case Analysis}

The smaller model's outputs frequently deviated from the requested JSON format in three characteristic ways:

\textbf{1. Markdown-Embedded JSON:} Instead of returning a clean JSON array, the model embedded JSON objects within markdown formatting:
\begin{verbatim}
1. **Summer Vacation (June-August)**
   - **Beach Relaxation**: Head to
     Florida for a beach vacation.
     - {"departure_airport": "IND",
       "arrival_airport": "PBI",
       "travel_probability": 0.15}
   - **Mountain Getaway**: Visit
     Colorado for outdoor activities.
     - {"departure_airport": "IND",
       ...
\end{verbatim}

\textbf{2. Natural Language Format:} Some responses used descriptive text rather than structured JSON:
\begin{verbatim}
1. Arrival Airport: MCO (Orlando)
   Travel Probability: 0.15
2. Arrival Airport: DEN (Denver)
   Travel Probability: 0.12
...
\end{verbatim}

\textbf{3. Truncated Responses:} With \texttt{max\_tokens=1024}, verbose responses were cut off mid-JSON, producing malformed output.

\subsection{Three-Tier Parsing Algorithm}

To address these failures, we implemented a three-tier parsing approach:

\textbf{Tier 1: Bracket Extraction (Original Method)}
\begin{enumerate}
\item Search for content between \texttt{[...]} brackets
\item Attempt JSON parsing with trailing comma handling
\item If successful, normalize probabilities and return
\end{enumerate}

\textbf{Tier 2: Regex JSON Extraction (New)}
\begin{enumerate}
\item If Tier 1 fails, apply regex pattern matching to find JSON-like objects containing airport fields embedded in the text
\item Extract all matching objects and validate that they contain required fields (departure airport, destination, probability)
\item Clean formatting issues: convert single quotes to double quotes, remove trailing commas
\item If successful, normalize probabilities and return
\end{enumerate}

\textbf{Tier 3: Natural Language Extraction (New)}
\begin{enumerate}
\item If Tier 2 fails, search for natural language patterns
\item Patterns include: ``Arrival: XXX'', ``Destination: Name (XXX)'', ``Travel Probability: 0.XX''
\item Extract airport codes and probability values via regex
\item Construct structured records from extracted data
\item If successful, normalize probabilities and return
\end{enumerate}

Additionally, \texttt{max\_tokens} was increased from 1024 to 4096 to prevent truncation of verbose responses.

\subsection{Original Results (Without Rectification)}
\label{sec:appendix_original_results}

For transparency, we present the original, smaller model results obtained with single-tier parsing. These results illustrate what researchers may encounter without specialized parsing for smaller models.

\emph{Original Persona Analysis Results:} Only 3/20 personas (15\%) were successfully parsed, yielding 27 unique destinations with 13.0 average destinations per persona. The remaining 17 personas returned malformed JSON.

\emph{Original Convergence Analysis Results:} Only 6/20 iterations (30\%) were parsed, producing 63 total records. Convergence analysis was not possible due to insufficient data.

These results highlight the importance of robust parsing strategies when working with smaller LLMs, which may produce semantically valid but syntactically non-compliant outputs.

\section{Detailed Experimental Results}
\label{sec:appendix_results}

\begin{figure*}[!htb]
\centering
\includegraphics[width=\textwidth]{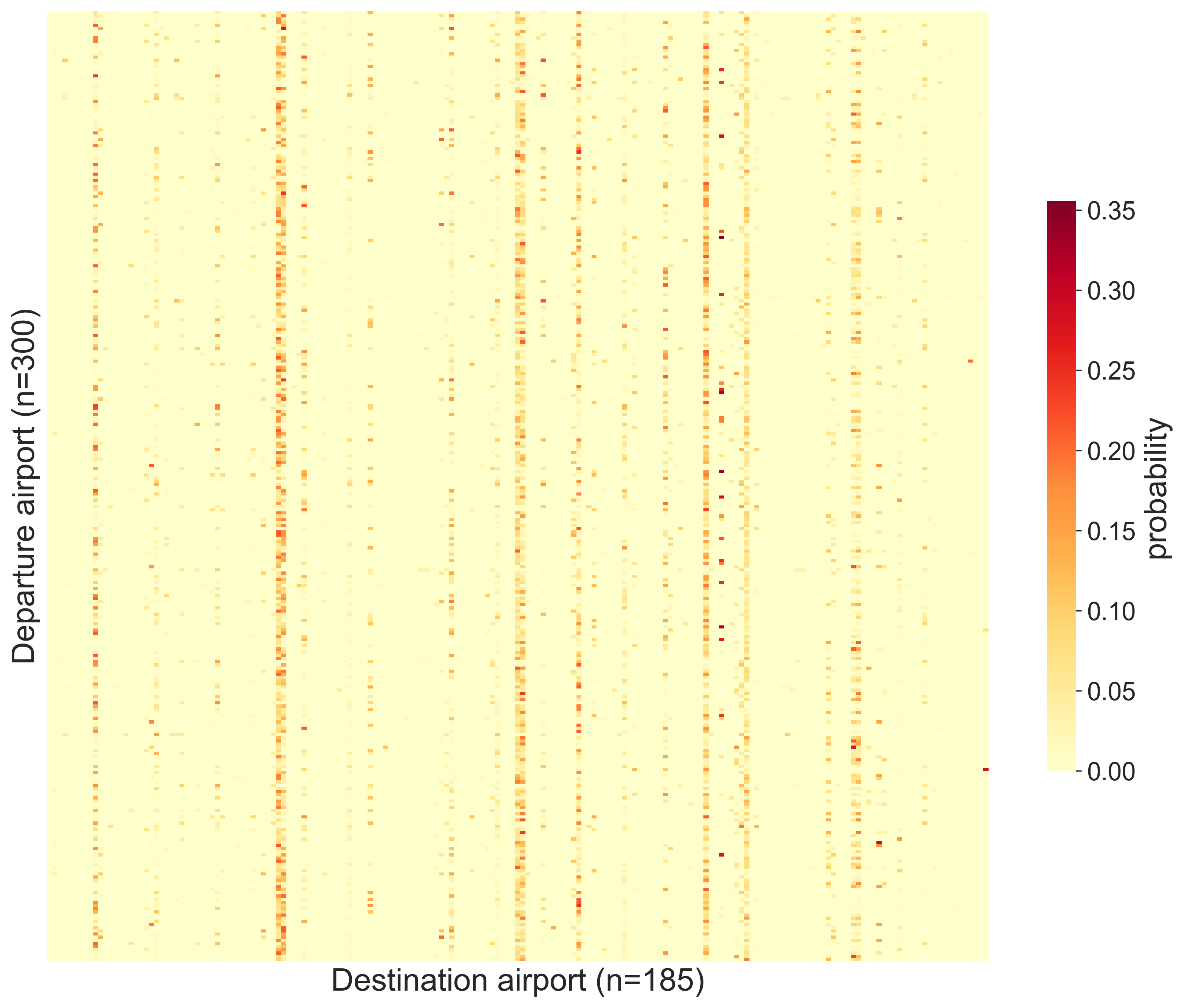}
\caption{Probability distribution heatmap showing 300 departure airports (rows) $\times$ 185 unique destination airports (columns) that appeared across all contexts. Of the 300 airports provided as options, only 185 received a non-zero probability in at least one context. Bright vertical bands reveal consistently high-probability modes.}
\label{fig:rq1_heatmap}
\end{figure*}

Figure~\ref{fig:rq1_heatmap} shows the complete probability matrix across 300 contexts, revealing that LLM-generated distributions encode clear geographic structure with major hub locations appearing as consistently high-probability modes.

\section{Convergence and Stability Analysis}
\label{sec:appendix_convergence}

\begin{figure*}[!htb]
\centering
\includegraphics[width=\textwidth]{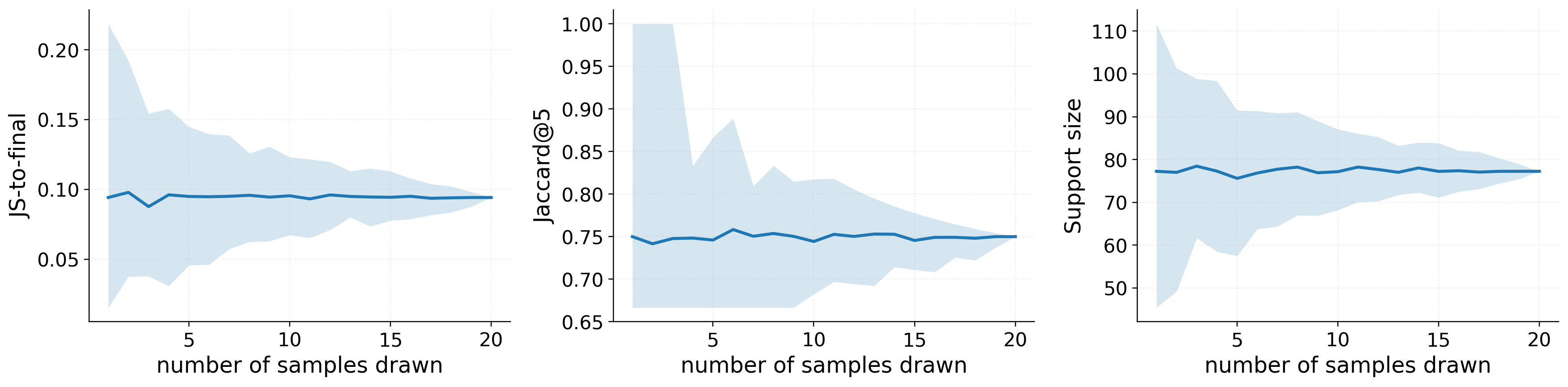}\\[4pt]
\includegraphics[width=\textwidth]{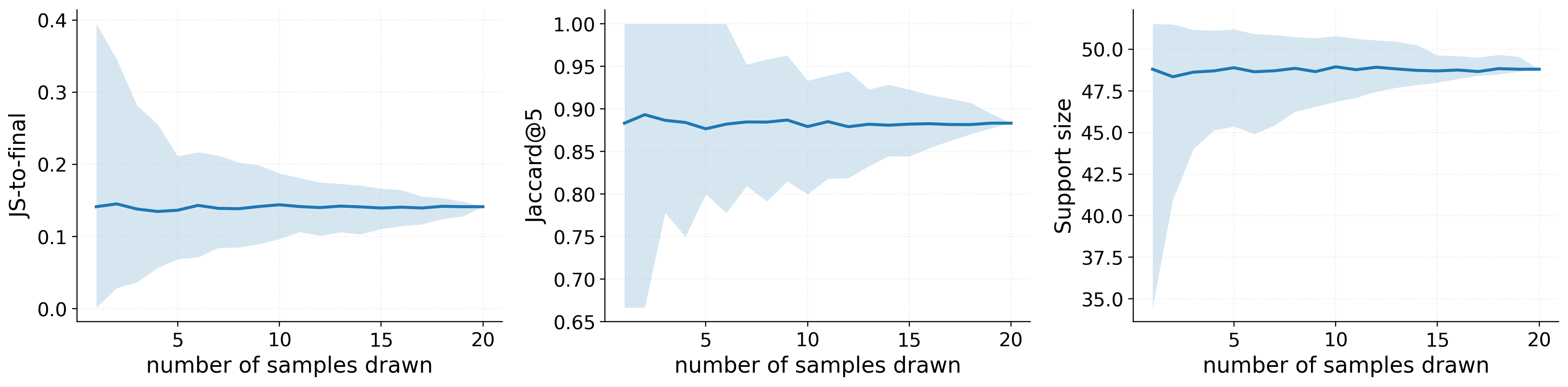}
\caption{Cross-domain convergence: Yelp (300 restaurants, top), Amazon (300 products, bottom). Resampling protocol: Section~\ref{sec:bootstrap_protocol}. Each panel plots the mean and $95\%$ band of JS-to-final, Jaccard@5 and support size against the number of samples drawn from the $20$-sample pool. The means are flat across the range while the bands contract, showing that both domains are already stable at small sample counts; the per-iteration convergence points are reported in Table~\ref{tab:cross_domain}.}
\label{fig:cross_domain}
\end{figure*}

\begin{figure*}[!htb]
\centering
\includegraphics[width=\textwidth]{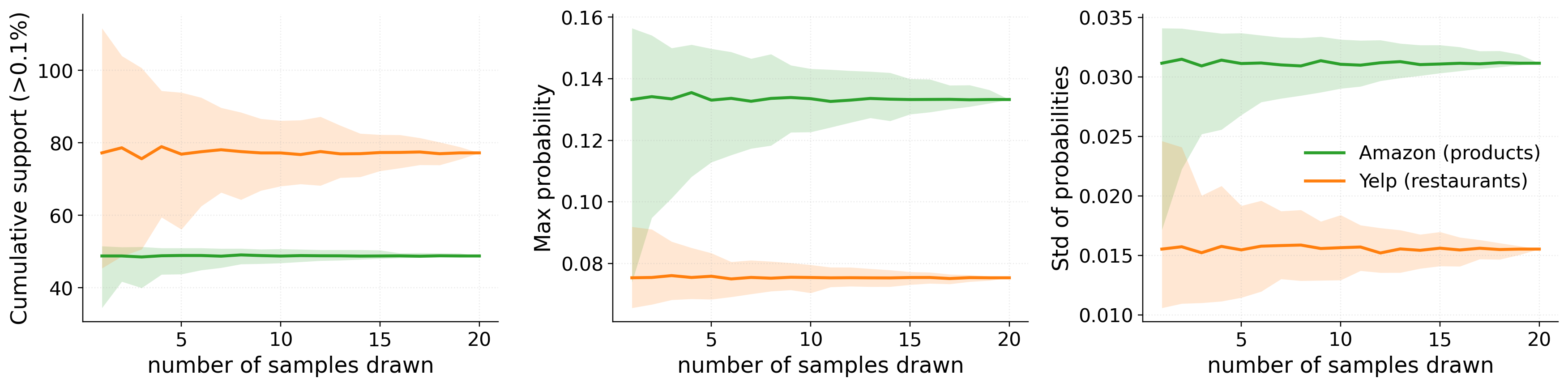}
\caption{Per-sample distributional signatures for the two cross-domain experiments (Yelp restaurants, Amazon products; Indianapolis, $20$ iterations; panels and resampling protocol as in Figure~\ref{fig:llm_convergence}). Cumulative support (panel~a) counts canonical items whose averaged, renormalized cumulative probability exceeds $0.1\%$, the same definition underlying the final-support values in Table~\ref{tab:cross_domain} ($103$ for Yelp, $51$ for Amazon at the full $20$-sample pool). Amazon concentrates probability on a few products (higher maximum probability, smaller support), whereas Yelp spreads mass across many restaurants, consistent with Amazon's higher top-10 mass ($79\%$) than Yelp's ($49\%$).}
\label{fig:cross_domain_signature}
\end{figure*}

Figure~\ref{fig:cross_domain} visualizes cross-domain convergence patterns discussed in Section~\ref{sec:cross_domain_results}, and Figure~\ref{fig:cross_domain_signature} shows the corresponding per-sample distributional signatures for the Yelp and Amazon domains, the cross-domain analogue of Figure~\ref{fig:llm_convergence}.

\begin{figure*}[!htb]
\centering
\includegraphics[width=\textwidth]{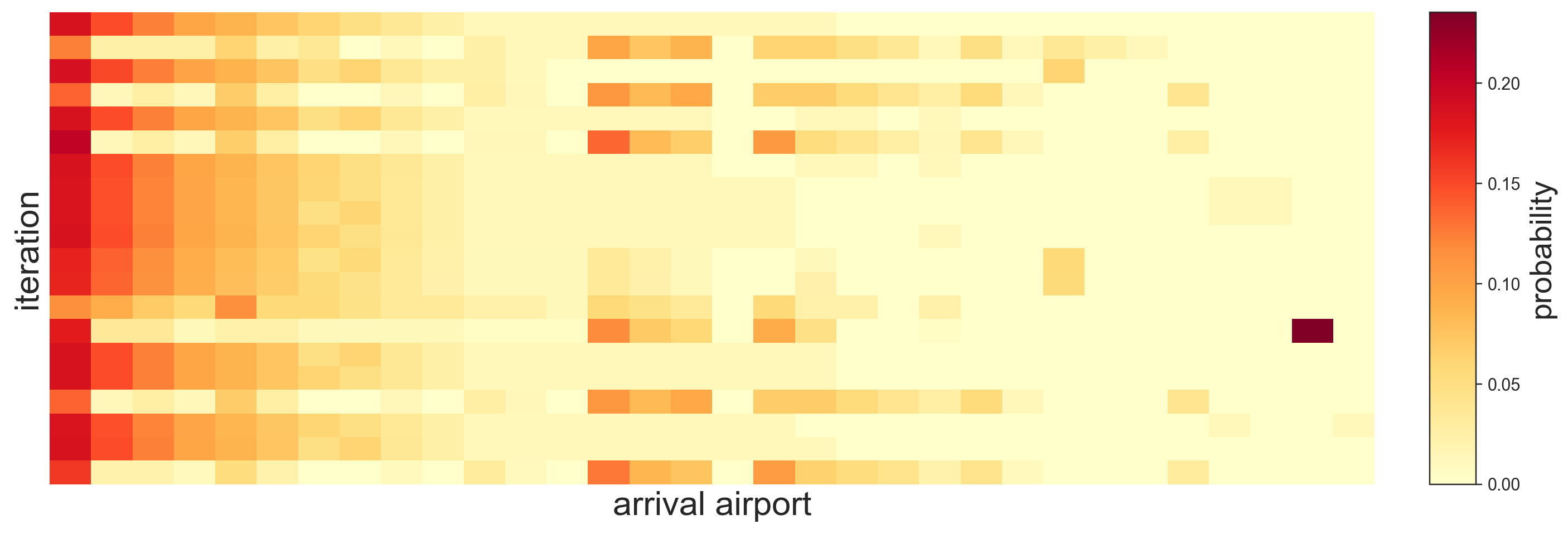}
\caption{Convergence heatmap for Regional Context A (IND) over 20 iterations. Rows: iterations (0--19). Columns: the distinct destinations observed across the run, i.e.\ this context's cumulative support (32 destinations). Bright vertical bands show stable high-probability modes; lighter peripheral colors show gradual tail expansion.}
\label{fig:rq3_ind}
\end{figure*}

\begin{figure*}[!htb]
\centering
\includegraphics[width=\textwidth]{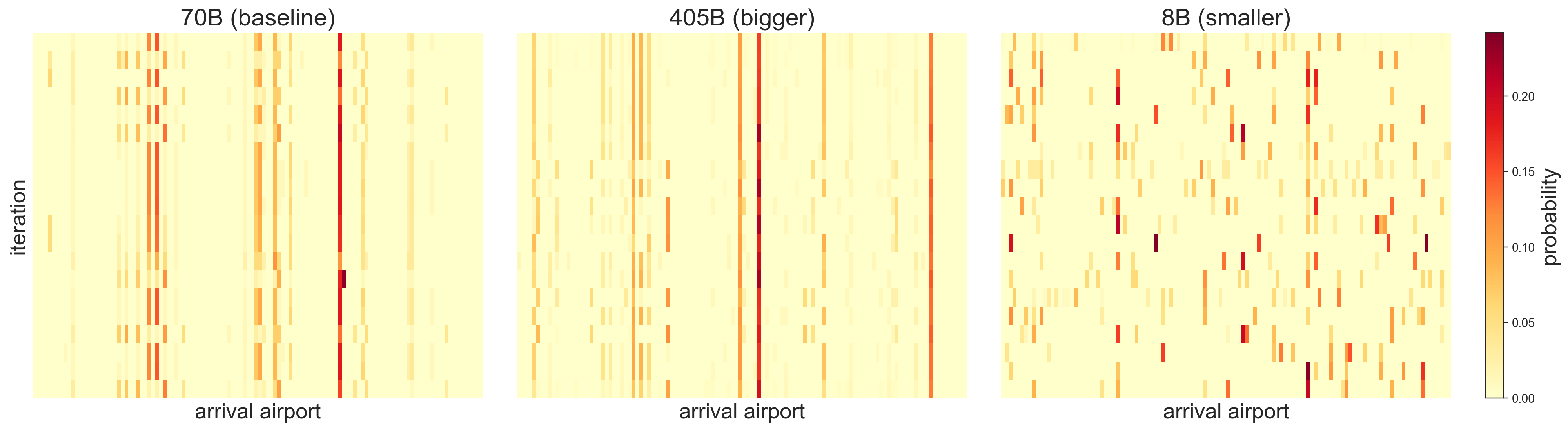}
\caption{Side-by-side comparison of convergence patterns across three LLM sizes with rectified parsing. Left: Baseline (70B, 32 final destinations); Center: Bigger (405B, 54 final destinations); Right: Smaller (8B, 100 final destinations). The smaller model achieves the highest diversity but requires more iterations for stabilization.}
\label{fig:exp3_llm_comparison}
\end{figure*}

Figure~\ref{fig:rq3_ind} provides a detailed iteration-by-iteration heatmap for a regional context (IND), demonstrating the modes-first, tails-later convergence pattern. Figure~\ref{fig:exp3_llm_comparison} compares this pattern side by side across the three model sizes.

\subsection{Extended Stability Analysis}
\label{sec:appendix_stability}

Per-scale and per-context stability metrics, namely JS divergence (overall and region-wise), Jaccard@$k$, and tail dynamics (support, entropy, tail mass), track the values reported in Tables~\ref{tab:convergence} and~\ref{tab:mode_tail_decomp}: pairwise divergence decreases and top-$k$ overlap increases with more samples, and the 8B model shows the highest run-to-run divergence (mean JS $=0.483$ vs.\ $0.171$ and $0.152$ for 70B and 405B) with a larger mode contribution to total divergence.

\section{Model Size Comparison: Full Heatmaps}
\label{sec:appendix_models}

\begin{figure}[t]
\centering
\includegraphics[width=\columnwidth]{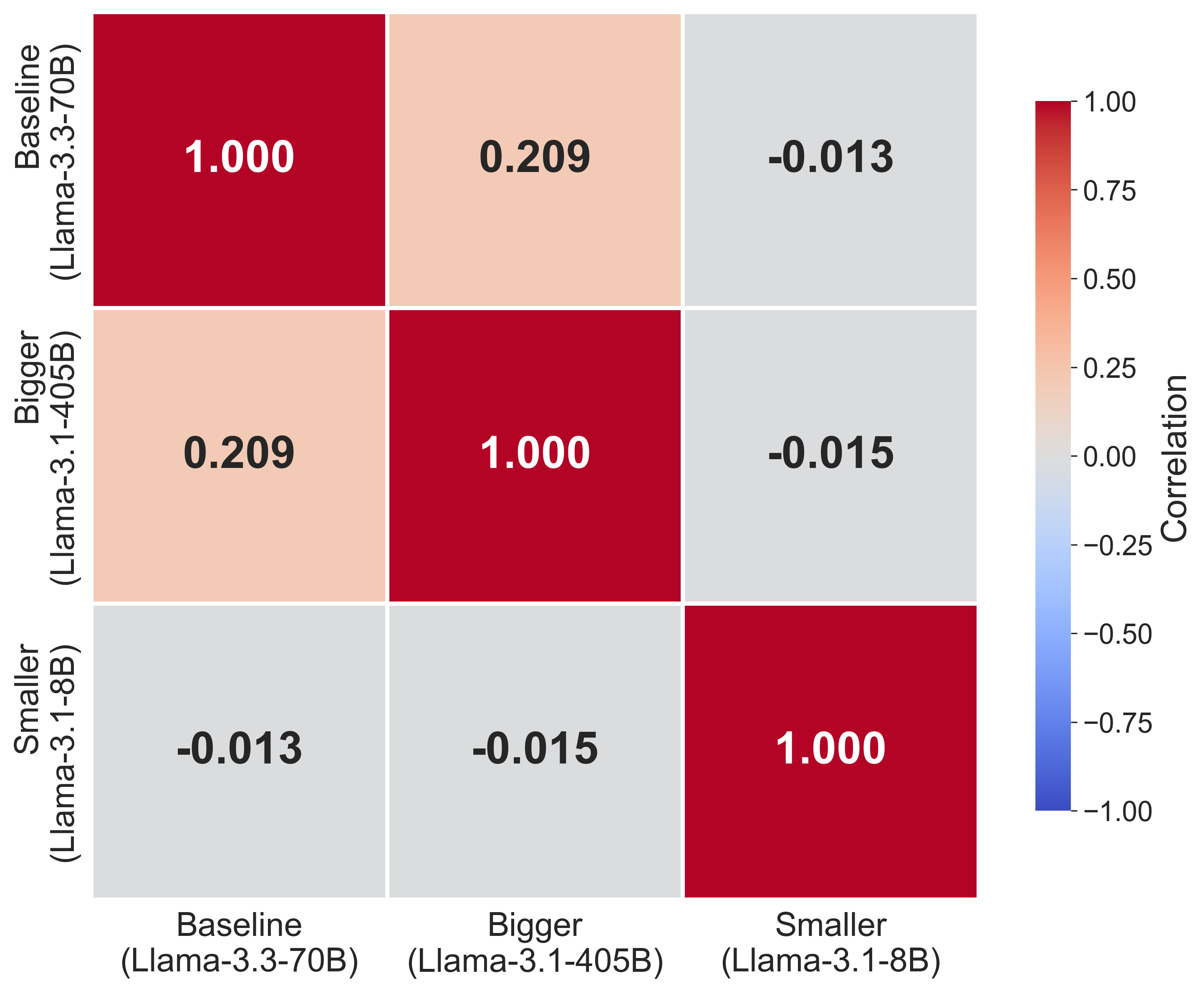}
\caption{Scale discordance matrix showing pairwise correlations between three LLM sizes (8B, 70B, 405B). Values near zero indicate minimal agreement, demonstrating that model scale does not induce smooth distributional refinement. Different models encode fundamentally different world priors.}
\label{fig:correlation_matrix}
\end{figure}

Figure~\ref{fig:correlation_matrix} visualizes the scale discordance phenomenon discussed in Section~\ref{subsec:scale_discordance}.

\begin{figure*}[!htb]
\centering
\includegraphics[width=\textwidth]{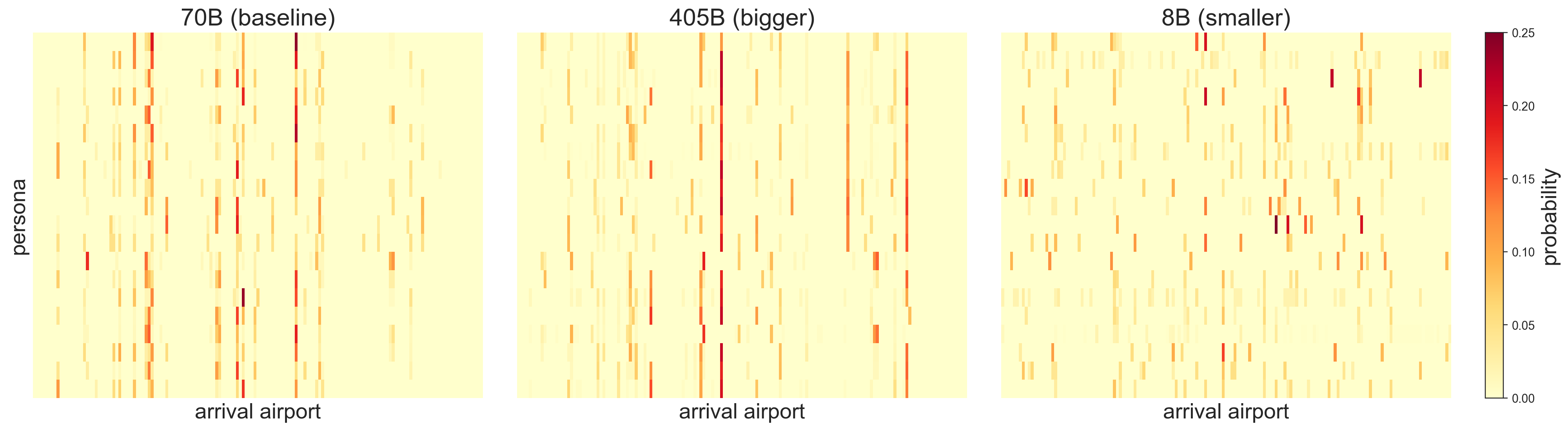}
\caption{Side-by-side comparison of persona-based distributions across three LLM sizes with rectified parsing. Left: Baseline (70B, 46 destinations); Center: Bigger (405B, 72 destinations); Right: Smaller (8B, 131 destinations). The smaller model produces the highest diversity when properly parsed.}
\label{fig:exp2_llm_comparison}
\end{figure*}

Figure~\ref{fig:exp2_llm_comparison} compares persona-based probability distributions side by side across the three model sizes.

\subsection{Non-Monotonic Scaling of Distributional Diversity}
\label{sec:appendix_nonmonotonic}

An intriguing observation emerges from the convergence analysis (Figure~\ref{fig:llm_convergence}a): the baseline model (70B) produces the fewest unique destinations at convergence (32), fewer than both the smaller 8B model (100) and the larger 405B model (54). These are cumulative counts at the full pool of $20$ samples, the right-hand end of each curve, whereas the caption of Figure~\ref{fig:llm_convergence} reports each curve's mean over the whole resampling range ($71.6$ for 8B, $45.2$ for 405B, and $29.3$ for 70B); both readings give the same ordering. This pattern reflects non-monotonic scaling in distributional diversity. The 8B model produces diffuse distributions with low mode concentration (max probability 9.0\%), spreading probability mass across many alternatives. Both 70B and 405B models achieve similar high mode concentration (17.1\% and 17.4\% respectively), but differ in tail behavior: the 70B model aggressively truncates low-probability alternatives, while the 405B model maintains a longer tail with more destinations receiving non-negligible probability. This suggests that medium-scale models may be optimized for decisiveness, forming strong preferences while pruning alternatives, whereas larger models balance confidence in top choices with greater distributional diversity. Such non-monotonic effects have implications for model selection: if tail diversity matters for an application, neither the smallest nor the largest model may be optimal.

\section{Cross-Family Methodology}
\label{sec:appendix_crossfamily}

This section documents the methodology for extending scale discordance analysis beyond the Llama family to include Qwen and Mistral, model families.

\subsection{Model Configuration}

We evaluate nine models spanning three families, each with small, medium, and large variants. The \emph{Llama family} (Meta, US) includes Llama-3.1-8B, Llama-3.3-70B, and Llama-3.1-405B, all using dense architectures. The \emph{Qwen family} (Alibaba, China) includes Qwen2.5-7B, Qwen2.5-72B (both dense), and Qwen3-235B, which uses a Mixture-of-Experts (MoE) architecture with 22B active parameters. The \emph{Mistral family} (Mistral AI, France) includes Mistral-7B-v0.3, Mistral-Small-24B (both dense), and Mistral-8x7B, which uses MoE with 12B active parameters out of 46B total. All models were accessed via the Together AI API.

\subsection{Per-Scale Distributional Signatures: Qwen and Mistral}
\label{sec:appendix_crossfamily_signatures}

\begin{figure*}[t]
\centering
\includegraphics[width=\textwidth]{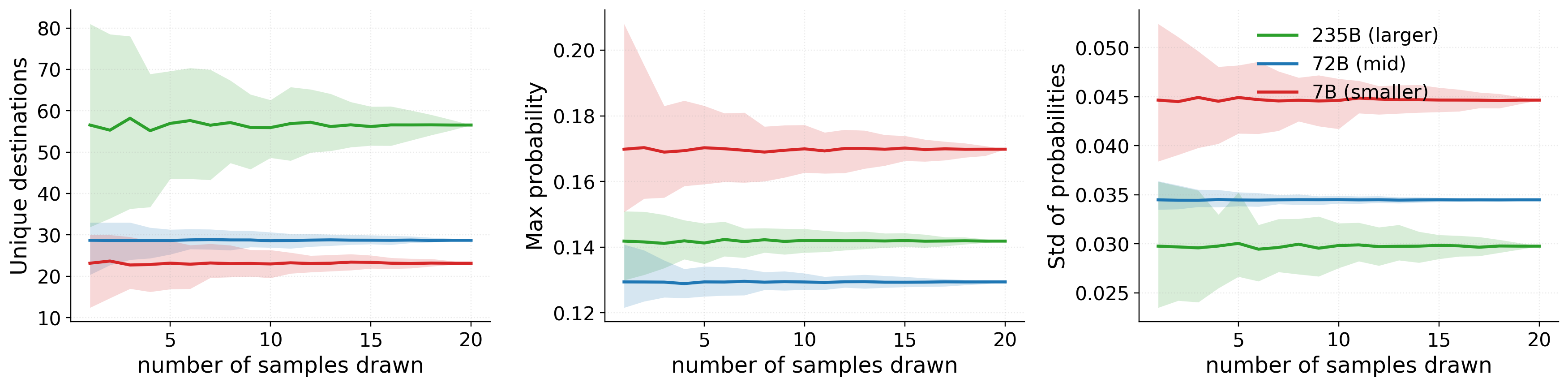}
\caption{Per-sample distributional signatures across three Qwen scales (Indianapolis, $20$ samples; Section~\ref{sec:bootstrap_protocol} resampling protocol; panels as in Figure~\ref{fig:llm_convergence}). The three scales occupy distinct operating points: the largest model (235B, a Mixture-of-Experts) is the most diffuse (highest unique-destination count), while the 7B and 72B models concentrate more probability mass on the top destinations.}
\label{fig:qwen_convergence}
\end{figure*}

\begin{figure*}[t]
\centering
\includegraphics[width=\textwidth]{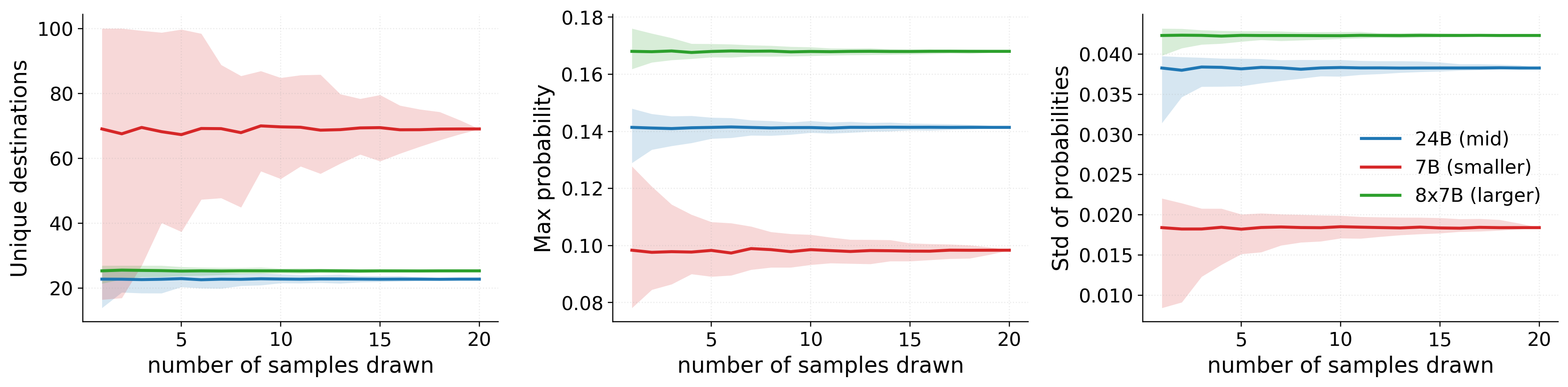}
\caption{Per-sample distributional signatures across three Mistral scales (Indianapolis, $20$ samples; Section~\ref{sec:bootstrap_protocol} resampling protocol; panels as in Figure~\ref{fig:llm_convergence}). The smallest model (Mistral-7B) is by far the most diffuse and least peaked, whereas Mistral-Small-24B and Mistral-8x7B concentrate mass on a few destinations, giving three distinct operating points rather than a refinement of one another.}
\label{fig:mistral_convergence}
\end{figure*}

Figures~\ref{fig:qwen_convergence} and~\ref{fig:mistral_convergence} give the per-scale distributional signatures for the Qwen and Mistral families (Indianapolis context, $20$ samples; the resampling protocol and the three panels match Figure~\ref{fig:llm_convergence} for Llama). The ordering of the separated operating points is family-dependent: the most diffuse scale is the \emph{smallest} in Llama and Mistral (8B, 7B) but the \emph{largest} in Qwen (235B), reinforcing the non-monotonic scaling of distributional diversity (Appendix~\ref{sec:appendix_nonmonotonic}).

\subsection{Output Normalization}
\label{sec:appendix_output_norm}

Models do not always emit destinations in the requested three-letter IATA format; we normalize these non-conforming outputs (informally, ``hallucinations'') before analysis, and discard any for which no valid code can be recovered. Different models produce destination outputs in varying formats, requiring normalization for cross-model comparison. While most models output standard three-letter IATA codes (e.g., ``ATL''), some models (notably Qwen3-235B and Mistral-Small-24B) output full airport names with embedded codes (e.g., ``Hartsfield Jackson Atlanta International Airport (ATL), Georgia'').

We implemented a normalization function that extracts three-letter IATA codes from any format using the following procedure:
\begin{enumerate}
    \item If the string is exactly three uppercase letters, return it directly
    \item Search for a three-letter uppercase code within parentheses using regex pattern \texttt{$\backslash$([A-Z]\{3\}$\backslash$)}
    \item If found, extract and return the matched code
\end{enumerate}

This normalization restored cross-model comparability. Without normalization, models using different output formats showed zero common destinations and maximum JS divergence (0.693). After normalization, typical common destination counts ranged from 15 to 30 between model pairs, and all cross-family scale pairs share substantial support; the reported JS divergences (Table~\ref{tab:cross_family_discordance}) are therefore well below the theoretical maximum of $\ln 2$, which is $0.693$ to three decimal places and would require completely non-overlapping supports.

\subsection{Mistral-Small-24B Behavioral Adaptation}

The Mistral-Small-24B model exhibited atypical behavior during data collection. Instead of generating travel demand distributions, it frequently requested clarification (e.g., ``Could you please specify the cities or regions where your scientific research projects are based?''). This behavior, characterized as RLHF over-alignment, resulted in an 11.2\% success rate (20 valid responses from 179 attempts).

Two mitigations were applied:
\begin{enumerate}
    \item \emph{Research context prefix:} Adding ``You are generating results for scientific research projects.'' to the prompt reduced clarification behavior by establishing a legitimate research context.
    \item \emph{Retry algorithm:} A retry mechanism with up to 500 attempts and incremental progress saving ensured the collection of the target 20 valid samples.
\end{enumerate}

This behavior was not observed in persona-conditioned experiments (Section~\ref{sec:appendix_personas}), suggesting that detailed demographic context naturally prevents clarification behavior by reducing prompt ambiguity.

\subsection{Experimental Parameters}

All cross-family experiments used consistent parameters matching the original Llama experiments:
\begin{itemize}
    \item \emph{Temperature:} 0.2 (low temperature for reproducibility)
    \item \emph{Iterations:} 20 per model (convergence analysis)
    \item \emph{Max tokens:} 4096 (increased from 1024 to accommodate verbose outputs)
    \item \emph{Departure context:} Indianapolis International Airport (IND)
    \item \emph{Choice set:} 300 US airports
\end{itemize}

All models were accessed via the Together AI API to ensure consistent infrastructure across families. Total data collection comprised 179 API calls for Mistral-Small-24B (due to retry requirements) and 20 calls each for the remaining eight models, yielding 20 valid samples per model for analysis.

\section{Persona-based Probability Distributions}
\label{sec:appendix_personas}

\begin{figure*}[!htb]
\centering
\includegraphics[width=\textwidth]{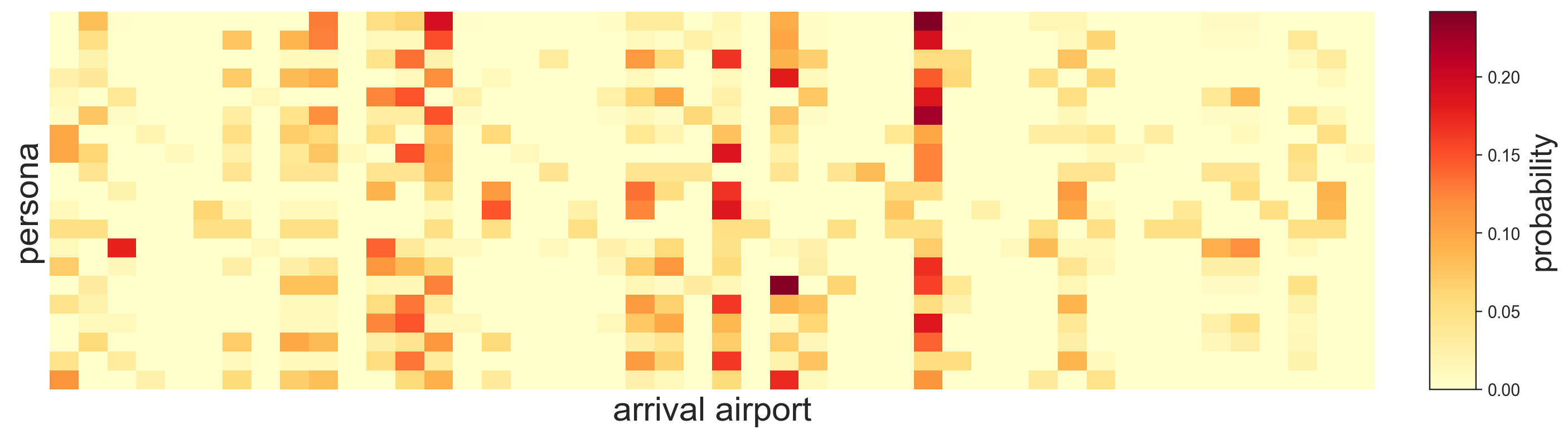}
\caption{Persona-based probability distributions from Regional Context A (IND). Rows: 20 demographic personas (Age $\times$ Education). Columns: destination outcomes. Personas aged 35 to 44 show the most concentrated preferences and senior personas the most diverse, by both peak probability and entropy.}
\label{fig:rq2_ind}
\end{figure*}

Figure~\ref{fig:rq2_ind} shows the complete persona-destination probability matrix for the Indianapolis context, revealing how demographic factors influence the full distributional structure.

\section{Extended Prompt Sensitivity Analysis}
\label{sec:appendix_sensitivity}

\subsection{Model Sensitivity vs Prompt Sensitivity}

Comparing model choice against prompt perturbations reveals that model selection has a larger impact on output distributions than persona conditioning.

\emph{Model Sensitivity:} Cross-model JS divergences are substantial: 70B vs 405B yields JS=0.365 (95\% CI: [0.33, 0.41]), 70B vs 8B yields JS=0.474 (95\% CI: [0.42, 0.53]), and 405B vs 8B yields JS=0.501 (95\% CI: [0.46, 0.54]).

\emph{Prompt Sensitivity:} In contrast, prompt perturbations produce smaller effects: persona conditioning yields JS=0.134 (95\% CI: [0.12, 0.14]), age perturbations yield JS=0.330 (95\% CI: [0.28, 0.38]), and education perturbations yield JS=0.261 (95\% CI: [0.23, 0.29]).

These results indicate that switching between model sizes (e.g., 8B to 70B) induces distributional shifts 2--4$\times$ larger than persona conditioning, and 1.1--1.9$\times$ larger than the demographic perturbations (age, education), highlighting model choice as a critical factor in LLM-based simulations.

\subsection{Seasonality Perturbation Details}
\label{sec:appendix_seasonality}

To further calibrate perturbation strength, we conduct a targeted seasonality test comparing summer vs winter travel contexts. We append a single sentence to the persona line specifying the travel season: ``You are planning travel for summer (June--August).'' vs.\ ``You are planning travel for winter (December--February).'', with the question correspondingly asking for travel plans ``for this season''. Using a fixed neutral persona and the baseline model (70B), we compute the JS distance between summer and winter distributions.

Table~\ref{tab:seasonality} reports the JS distance between summer and winter distributions across four departure airports representing different geographic contexts and sizes. Because that table is on the distance scale and Table~\ref{tab:prompt_effects} on the divergence scale (Appendix~\ref{sec:appendix_metrics}), the two must be placed on a common scale before they are compared. Squaring each airport value and averaging gives a mean seasonal divergence of $0.098$, below every perturbation in Table~\ref{tab:prompt_effects}: age ($0.33$), context ($0.31$), education ($0.26$), and persona ($0.13$). The ordering does not depend on the scale, since in distance units the seasonal effect is $0.311$ against $0.574$ for age and $0.511$ for education. We report this as an ordering of point estimates rather than a separation, since no confidence interval is estimated for the four-airport mean. Notably, large hub airports (ATL, ORD) show smaller seasonal effects than smaller airports (FWA, IND), suggesting that major hubs serve as stronger geographic anchors that partially buffer against temporal context changes.

\begin{table}[t]
\centering
\small
\setlength{\tabcolsep}{4pt}
\begin{tabular}{@{}lcc@{}}
\toprule
\textbf{Airport} & \textbf{JS Dist.} & \textbf{95\% CI} \\
\midrule
FWA (Fort Wayne) & 0.359 & [0.316, 0.412] \\
IND (Indianapolis) & 0.334 & [0.298, 0.375] \\
ATL (Atlanta) & 0.285 & [0.255, 0.317] \\
ORD (Chicago) & 0.264 & [0.230, 0.306] \\
\midrule
\textbf{Mean} & \textbf{0.311} & --- \\
\bottomrule
\end{tabular}
\caption{
Seasonality perturbation effect sizes for travel demand (US airports). JS distance (the square root of the JS divergence; Appendix~\ref{sec:appendix_metrics}) between summer (June--August) and winter (December--February) distributions. Smaller airports show larger seasonal effects than major hubs.}
\label{tab:seasonality}
\end{table}

\section{Temperature Sensitivity Analysis: Extended Results}
\label{sec:appendix_temperature}

\begin{table}[h]
\centering
\small
\setlength{\tabcolsep}{4pt}
\begin{tabular}{@{}c|cccccc@{}}
\toprule
$T$ & Unique & Max P & Entropy & Top-5 & Jac@5 & Jac@10 \\
\midrule
0.1 & 30.5 & 0.176 & 3.93 & 0.573 & 0.924 & 0.885 \\
\textbf{0.2} & \textbf{33.2} & \textbf{0.164} & \textbf{4.07} & \textbf{0.540} & \textbf{0.929} & \textbf{0.911} \\
0.3 & 37.5 & 0.160 & 4.16 & 0.530 & 0.924 & 0.853 \\
0.4 & 33.3 & 0.172 & 4.07 & 0.550 & 0.883 & 0.886 \\
0.5 & 36.7 & 0.169 & 4.14 & 0.541 & 0.901 & 0.890 \\
0.6 & 35.7 & 0.175 & 4.08 & 0.551 & 0.873 & 0.886 \\
0.7 & 36.8 & 0.172 & 4.13 & 0.539 & 0.853 & 0.887 \\
0.8 & 41.3 & 0.165 & 4.24 & 0.519 & 0.766 & 0.865 \\
0.9 & 40.3 & 0.151 & 4.35 & 0.494 & 0.834 & 0.814 \\
1.0 & 41.8 & 0.149 & 4.34 & 0.494 & 0.813 & 0.901 \\
\bottomrule
\end{tabular}
\caption{Temperature effects on convergence metrics (mean across 6 settings; Unique, Max P, Entropy and Top-5 at the final iteration, Jac@5 and Jac@10 averaged over all iterations). Bold indicates the primary experimental setting ($T=0.2$). All ten sampled temperatures are listed. Jac@10 is reported alongside Jac@5 to show that mode-set stability is preserved at both cutoffs across the full temperature range, so the mode--tail threshold of Appendix~\ref{sec:appendix_k_justification} is not specific to $T{=}0.2$. Jac@5 declines overall but not monotonically, rising at $T{=}0.1 \rightarrow 0.2$, at $T{=}0.4 \rightarrow 0.5$, and at $T{=}0.8 \rightarrow 0.9$; the first of these is what makes $T{=}0.2$ the highest Jac@5 in the sweep.}
\label{tab:temperature_summary}
\end{table}

This appendix provides detailed results from the temperature sensitivity analysis described in Section~\ref{sec:temperature_sensitivity}.

\subsection{Experimental Design}
\begin{figure*}[t]
\centering
\includegraphics[width=\textwidth]{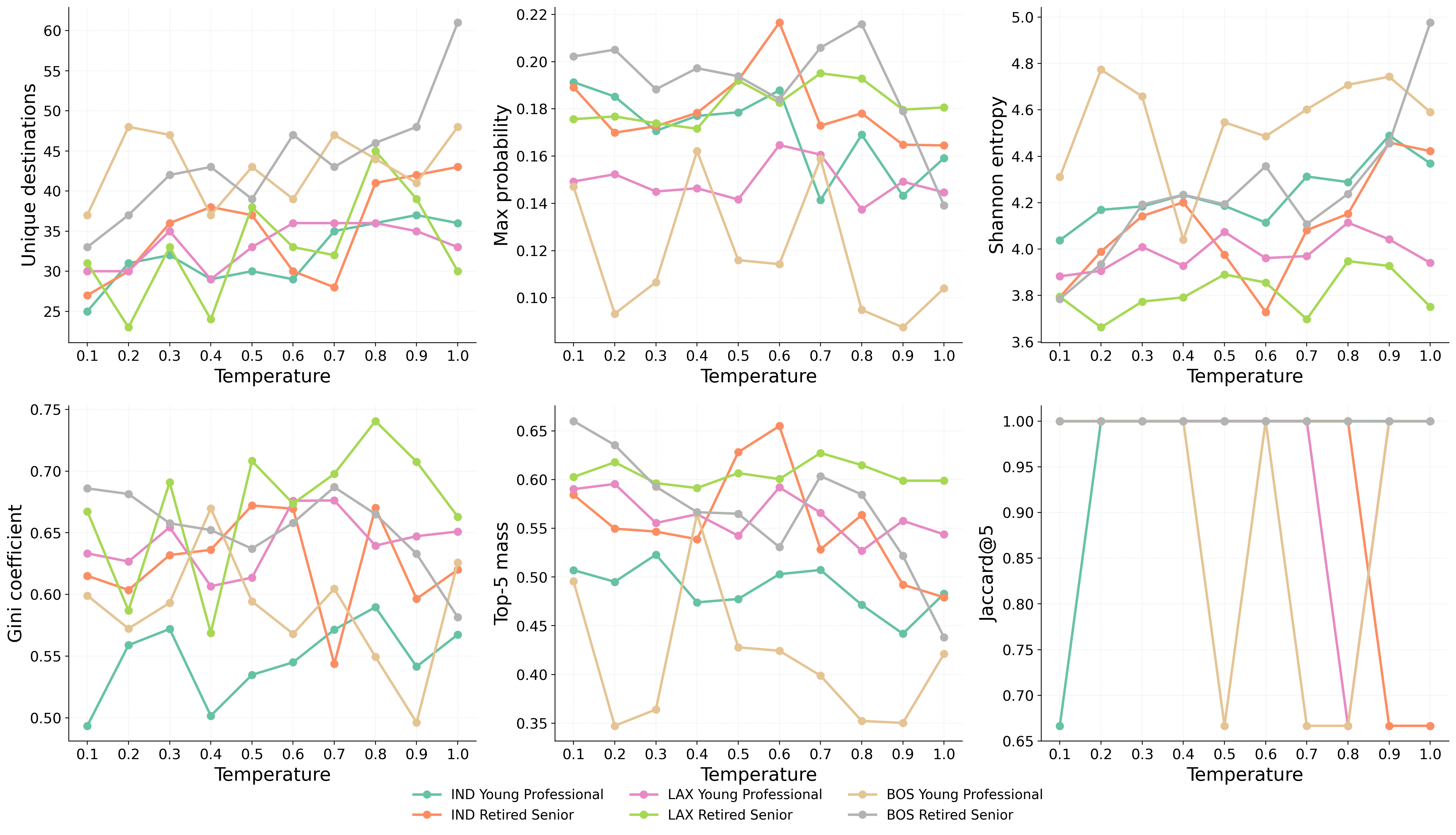}
\caption{Temperature effects across 6 experimental settings (3 airports $\times$ 2 personas) at the final iteration. Each panel shows a different metric plotted against temperature ($T \in \{0.1, \ldots, 1.0\}$). Lines represent different setting combinations. Key observations: (1) Unique destinations increase with temperature across most settings; (2) Shannon entropy consistently increases with temperature (6/6 settings); (3) Top-5 mass consistently decreases with temperature (6/6 settings); (4) Jaccard@5 stability peaks at low temperatures and degrades at high temperatures. These patterns confirm that temperature effects are systematic and generalizable across diverse experimental configurations.}
\label{fig:temperature_effects}
\end{figure*}

We tested 10 temperature values ($T \in \{0.1, 0.2, 0.3, 0.4, 0.5, 0.6, 0.7, 0.8, 0.9, 1.0\}$) across 6 settings (3 airports $\times$ 2 personas) with 10 iterations per combination, totaling 600 API calls. Table~\ref{tab:temperature_summary} summarizes the resulting metrics, and Figure~\ref{fig:temperature_effects} plots them across settings.

\subsection{Temperature Summary Table}

Table~\ref{tab:temperature_summary} reports the cross-setting means for every sampled temperature, giving the per-temperature view behind the aggregate statistics summarized below.

\subsection{Statistical Tests}
Kruskal-Wallis ANOVA confirms significant temperature effects for max probability (6/6 settings, mean $\eta^2 = 0.63$), entropy (6/6 settings, mean $\eta^2 = 0.45$), and top-5 mass (6/6 settings, mean $\eta^2 = 0.60$); all 18 tests reach $p < 0.001$, the weakest being $p = 5.8 \times 10^{-4}$. Low-versus-high temperature contrasts on these three metrics are large in 13 of 18 comparisons, with $\lvert d\rvert$ ranging from 0.98 to 3.50. The cross-setting averages move only modestly and non-monotonically across the whole sampled range, reversing at least once above $T{=}0.7$ as well as in the mid-range, with the clearest shifts at the extremes; the test is significant because the \emph{direction} of change is nonetheless consistent across settings as $T$ rises, with entropy increasing in 6/6 settings and top-5 mass decreasing in 6/6. Ranking stability declines in 5/6 settings (significantly in 4/6); the exception is flat rather than reversed (Spearman $r = +0.065$, $p = 0.54$). Jaccard@5 is discrete, admitting only six values on size-$5$ sets (Appendix~\ref{sec:appendix_stab_threshold}) and taking four of them across these runs, so per-setting trend tests have limited power; the aggregate decline is unambiguous ($r = -0.930$ across the ten cross-setting means, $p = 0.0001$).

\section{Greedy Decoding ($T{=}0$): Extended Results}
\label{sec:appendix_greedy}

\begin{table}[h]
\centering
\small
\setlength{\tabcolsep}{5pt}
\begin{tabular}{@{}llcc@{}}
\toprule
\textbf{Metric (Claim)} & & \textbf{$T{=}0.2$} & \textbf{$T{=}0$} \\
\midrule
\multicolumn{4}{l}{\textit{Convergence}} \\
\quad IND unique dest, iter 20 & & 32 & 27 \\
\quad IND stabilization & & Iter 3 & Iter 2 \\
\midrule
\multicolumn{4}{l}{\textit{Scale Discordance}} \\
\quad 8B unique dest, iter 20 & & 100 & 100 \\
\quad 405B stabilization & & Iter 4 & Iter 2 \\
\midrule
\multicolumn{4}{l}{\textit{Bounded Perturbations}} \\
\quad Persona avg dest/persona & & 17.8 & 17.6 \\
\quad Seasonality mean JS dist. & & 0.311 & 0.354 \\
\bottomrule
\end{tabular}
\caption{Comparison of key metrics between stochastic ($T{=}0.2$) and greedy ($T{=}0$) decoding across the three core claims. All qualitative findings are preserved.}
\label{tab:greedy_comparison}
\end{table}

\subsection{Experimental Design}

All experiments in this appendix replicate the corresponding $T{=}0.2$ experiments with identical configurations (models, prompts, iteration counts, and evaluation metrics) except that the sampling temperature is set to $T{=}0$ (greedy decoding). This eliminates all sampling stochasticity, isolating the deterministic component of model behavior.

Table~\ref{tab:greedy_comparison} compares key metrics across the three core claims. The modes-first convergence pattern holds at $T{=}0$, with faster stabilization (Iter~2 vs.\ Iter~3 for IND) and slightly fewer tail destinations (27 vs.\ 32), consistent with removing sampling noise that primarily affects low-probability outcomes. Residual non-determinism persists even at $T{=}0$ due to GPU floating-point effects, confirming that the convergence phenomenon is not purely a sampling artifact. Scale discordance persists and intensifies: the 8B model still produces 100 unique destinations with diffuse probability mass, and inter-model correlations remain near zero ($r=0.056$). Bounded perturbation effects are nearly identical: persona-driven destination counts average 17.6 at $T{=}0$ versus 17.8 at $T{=}0.2$, while seasonality-induced distributional shifts persist (JS distance $=0.354$).

Per-condition breakdowns (convergence, multi-scale, cross-family, persona, cross-domain, and seasonality at $T{=}0$) reproduce the corresponding $T{=}0.2$ results summarized above and are omitted here for brevity.
\section{Statistical Robustness: Extended Results}
\label{sec:appendix_robustness}

This appendix provides detailed results from the multi-source robustness analysis described in Section~\ref{sec:statistical_robustness}.

\subsection{Three Robustness Sources}
We test three orthogonal sources of variability. \emph{Source A} (permutation) uses 15 random orderings of the airport list with 15 iterations each (225 calls). \emph{Source B} (paraphrase) uses 5 semantically equivalent prompt variants with 15 iterations each (75 calls). \emph{Source C} (origins) uses 3 departure airports (IND, LAX, BOS) with 15 iterations each (45 calls).

\subsection{Coefficient of Variation by Source}

Table~\ref{tab:robustness_cv} reports the coefficient of variation of each metric under the three sources, so their relative contributions can be compared directly.

\begin{table}[h]
\centering
\small
\begin{tabular}{@{}l|ccc@{}}
\toprule
Metric & CV Perm. & CV Para. & CV Origins \\
\midrule
Unique Dest. & 13.2\% & 14.6\% & 20.5\% \\
Max Prob. & 10.1\% & 13.6\% & 13.3\% \\
Entropy & 5.1\% & 4.6\% & 5.4\% \\
Top-5 Mass & 10.1\% & 9.1\% & 10.0\% \\
\bottomrule
\end{tabular}
\caption{Coefficient of variation across three robustness sources. All values below 21\% indicate good reproducibility.}
\label{tab:robustness_cv}
\end{table}

\subsection{Variance Decomposition (ICC)}

Table~\ref{tab:icc_variance} reports an intraclass-correlation decomposition, ICC(1,1), of each metric under three grouping factors: departure airport, persona, and model scale. This isolates how much of the total variance each factor accounts for, complementing the pairwise divergences of Section~\ref{subsec:scale_discordance}.

\begin{table}[h]
\centering
\small
\begin{tabular}{@{}l|ccc@{}}
\toprule
Metric & Airport & Persona & Model \\
\midrule
Max Prob. & $-0.02$ & $0.16$ & $0.20$ \\
Entropy & $-0.02$ & $-0.40$ & $0.17$ \\
Top-5 Mass & $0.08$ & $-0.07$ & $-0.01$ \\
Support & $0.07$ & $-0.31$ & $\mathbf{0.46}$ \\
\bottomrule
\end{tabular}
\caption{ICC(1,1) variance decomposition by grouping factor. A negative value indicates that within-group variance exceeds between-group variance, so the factor accounts for no systematic share. Departure airport stays at or below $0.08$ on every metric and persona is at or below zero except on peak probability, whereas model scale is the only factor with a non-trivial positive component and is largest on distributional support.}
\label{tab:icc_variance}
\end{table}

\subsection{Consensus Destinations}
Top destinations appearing consistently across 15 input orderings: ORD (15/15 top-5), DFW (14/15), MCO (12/15), LAX (11/15), DEN (10/15).

\end{document}